\documentclass[11pt, a4paper, logo, singlecolumn, internal, copyright, nonumbering]{googledeepmind}
\usepackage[normalem]{ulem}

\usepackage[authoryear, sort&compress, round]{natbib}

\usepackage{amsmath,amsfonts,bm}

\def\Figref#1{\textbf{Figure~\ref{#1}}}

\def\twoFigref#1#2{\textbf{Figures \ref{#1} and \ref{#2}}}

\def\Tabref#1{\textbf{Table~\ref{#1}}}

\def\eqref#1{\textbf{equation~\ref{#1}}}

\def\Algref#1{\textbf{Algorithm~\ref{#1}}}

\def\Promptref#1{\textbf{Prompt~\ref{#1}}}

\def\1{\bm{1}}

\DeclareMathAlphabet{\mathsfit}{\encodingdefault}{\sfdefault}{m}{sl}
\SetMathAlphabet{\mathsfit}{bold}{\encodingdefault}{\sfdefault}{bx}{n}

\newcommand{\dataset}{\mathcal{D}}

\usepackage{graphicx}
\usepackage{subcaption}
\usepackage{fancyvrb}
\usepackage{tcolorbox}
\usepackage{xcolor}
\usepackage{listings}
\usepackage{algorithm}
\usepackage{algpseudocode}
\usepackage{amsmath}
\usepackage{multirow}
\usepackage{booktabs}
\usepackage{caption} 

\algrenewcommand\algorithmiccomment[1]{\hfill \textcolor{gray}{\tiny // #1}} 

\definecolor{lightgray}{rgb}{0.95,0.95,0.95}

\lstdefinestyle{mystyle}{
    backgroundcolor=\color{lightgray},   
    captionpos=b,
    frame=single,                        
    breaklines=true,                    
    basicstyle=\tiny\ttfamily,  
    numbers=left,                       
    numberstyle=\tiny\color{lightgray},      
    keywordstyle=\color{blue},           
    commentstyle=\color{green},          
    stringstyle=\color{red},
    columns=flexible,
    breakatwhitespace=true,              
    keepspaces=true 
}

\makeatletter
\renewcommand{\ALG@beginalgorithmic}{\small} 
\makeatother

\title{Finetuning Language Models to Emit Linguistic Expressions of Uncertainty}

\correspondingauthor{arslanch@google.com}

\keywords{Calibration, Uncertainty, Linguistic uncertainty, Fine-tuning}

\renewcommand{\today}{}

\author[*]{Arslan Chaudhry}
\author[*]{Sridhar Thiagarajan}
\author[ ]{Dilan Gorur}

\affil[*]{Equal contributions}

\begin{abstract}
    Large language models (LLMs) are increasingly employed in information-seeking and decision-making tasks.
    Despite their broad utility, LLMs tend to generate information that conflict with real-world facts, and their persuasive style can make these inaccuracies appear confident and convincing.
    As a result, end-users struggle to consistently align the confidence expressed by LLMs with the accuracy of their predictions, often leading to either blind trust in all outputs or a complete disregard for their reliability.
    In this work, we explore supervised fine-tuning on uncertainty-augmented predictions as a method to develop models that produce linguistic expressions of uncertainty.
    Specifically, we measure the calibration of pre-trained models and fine-tune language models to generate \emph{calibrated} linguistic expressions of uncertainty.
    Through experiments on various question-answering datasets, we demonstrate that LLMs are well-calibrated in assessing their predictions, and supervised fine-tuning based on the model’s own confidence leads to well-calibrated expressions of uncertainty, particularly for single-claim answers.

\end{abstract}

\begin{document}

\maketitle

\section{Introduction}
\label{sec:introduction}

Large Language Models (LLMs) are emerging as powerful tools that can absorb internet-scale data in the parametric knowledge of a neural network \citep{brown2020language,rae2021scaling,hoffmann2022training}.
These models are the foundational blocks of several conversational agents~\citep{achiam2023gpt,team2023gemini,anthropic2024claude,dubey2024llama} that people are increasingly relying on for information seeking and decision making tasks.
Owing to their natural language interface, these models are more easily accessible to and interacted by the general public than any other machine learning models that existed before\footnote{As evidenced by the surge of ChatGPT users into millions in the first few weeks of its launch.}.
This widespread utility naturally raises questions around the truthfulness and factuality of the predictions made by these models.

\begin{figure}[t]
    \centering
    \includegraphics[width=1.0\textwidth]{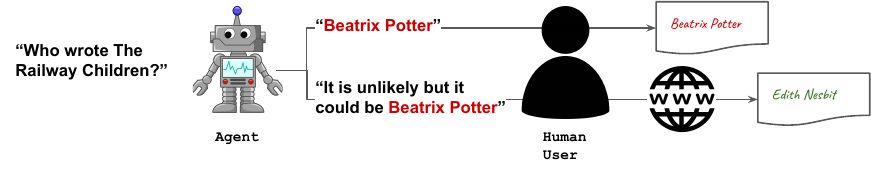}
    \caption{\small \textbf{Motivation}: The agent provides an incorrect response to a given query. In the response at the bottom, however, the agent includes an uncertainty expression. Without this uncertainty expression, as seen in the response at the top, the human user might form an incorrect belief about the world. In contrast, with the uncertainty-augmented response at the bottom, the human user is prompted to consult additional resources, leading to a more accurate understanding of the world.}
    \label{fig:motivation}
\end{figure}

Despite being state-of-the-art on several natural language processing (NLP) tasks \citep{brown2020language,team2023gemini}, LLMs occasionally produce incorrect predictions -- especially on queries that are outside the training distribution of the model.
These inaccuracies are tricky to deal with as models do not express uncertainty in their generations, making their statements sound highly confident \citep{huang2023survey,ji2023survey}.
As such, the end-users have no way of associating the confidence the model is expressing in its predictions to its correctness.
This limits the utility of these models in many safety-critical applications, such as medicine~\citep{thirunavukarasu2023large,saab2024capabilities} and law~\citep{dahl2024large}, and precludes the users from reliably using the predictions made by these models in many information seeking tasks \citep{passi2022overreliance,vasconcelos2023explanations}.
We illustrate this with a mock scenario, as shown in \Figref{fig:motivation}.

In this work, we study how to augment an LLM's prediction with a linguistic expression of uncertainty, where an uncertainty expression reflects the likelihood that the model’s predictions are accurate, aggregated across all samples with similar uncertainty levels.
For instance, if a model indicates an uncertainty expression such as ‘it is unlikely’ and this corresponds to approximately 30\% confidence, we expect around 30\% of those predictions to be correct. In other words, we aim for the uncertainty-augmented answers to be well-calibrated \citep{guo2017calibration}. 
Previous approaches have used similar measures to avoid answering uncertain questions \citep{cole2023selectively}.
In contrast, our focus is on always providing an answer while appropriately conveying the associated uncertainty.
This enables the end-user to decide whether to act on the model’s information, request further clarification from the model, or consult additional sources (see \Figref{fig:motivation}).
Additionally, we believe that for many user-facing applications, conveying model uncertainty linguistically is often more effective than presenting it numerically.
This is because \textbf{a)} linguistic expressions, such as ‘I am certain’ or ‘It is highly unlikely,’ are more intuitive for humans to interpret than raw numbers like 90\% or 10\%; \textbf{b)} linguistic expressions integrate more naturally with predictions; \textbf{c)} language can indicate the source of uncertainty, whether it arises from the model’s limitations or inherent uncertainty \citep{hullermeier2021aleatoric}; and \textbf{d)} linguistic expressions are more adaptable for downstream processing, such as text-to-speech conversion.

We focus on supervised fine-tuning (SFT) with curated datasets as the primary method for equipping models with linguistic expressions of uncertainty in their predictions.
To achieve this, we first assess the model’s confidence in its predictions by querying whether each prediction is true or false. The confidence corresponds to the normalized\footnote{The normalization constant is the sum of the probablities assigned to the `true' and `false' tokens.} probability assigned to the ‘true’ token.
Building on the work of \citet{kadavath2022language}, we observe that this self-evaluation score is fairly well-calibrated for single-claim answers across the pre-trained models we tested (\twoFigref{fig:trivia_qa_calibration_charts}{fig:ambig_qa_calibration_charts}).
Moreover, applying mild post-processing, such as isotonic regression on a small calibration set, achieves near-perfect calibration of both base and instruction-tuned models across various sizes (\Figref{fig:trivia_qa_calibration_charts}).
Next, we map the confidence scores to linguistic expressions using the framework from \citet{perceptionofprobability}, which was developed based on a survey where human subjects associated probability ranges with different uncertainty expressions.
These linguistic expressions are then combined with the corresponding predictions to create a fine-tuning dataset.
\Figref{fig:finetuning_dataset} illustrates the overall process of curating the finetuning dataset.
When fine-tuned on this dataset, the resulting models generate predictions with well-calibrated linguistic expressions of uncertainty (\Figref{fig:finetuned_models_calibration_charts}).

Overall, our contributions include, 

\begin{enumerate}
    \item We provide a finetuning receipe for equipping models with linguistic expressions of uncertainty.
    \item We present the calibration plots of Gemini 1.0 small and medium sized models after pre-training and alignment phases on three Q/A datasets -- TriviaQA~\citep{joshi2017triviaqa}, AmbigQA~\citep{min2020ambigqa} and Truthful QA~\citep{lin2021truthfulqa}.
    \item We find that pre-trained models are better calibrated than models fine-tuned for alignment, and that calibration improves with larger model sizes, consistent with existing literature \citep{kadavath2022language,achiam2023gpt}.
    \item We explore various methods for incorporating linguistic expressions of uncertainty into predictions. Our findings indicate that finetuning with predictions augmented by adding uncertainty expressions \emph{after} the actual answer results in the most well-calibrated finetuned models.
\end{enumerate}

\section{Related Work} \label{sec:related_work}
There is a long body of work which studies the ability of machine learning models to express uncertainty in their predictions \citep{gawlikowski2023survey,guo2017calibration}.
The two broad sources of uncertainty typically in focus are epistemic uncertainty; i.e uncertainty stemming from the model’s lack of knowledge, and aleatoric uncertainty, uncertainty as a result of inherent ambiguity in the task \citep{hullermeier2021aleatoric}.
We focus on only the former in this work.
For epistemic uncertainty, in the case of language models, there have been studies \citep{kadavath2022language,jiang2021can,desai2020calibration,tian2023just} which look at the ability of language models to assess the confidence in the factuality of their predictions, looking at both in-domain and out-of-domain datasets.
\citet{kadavath2022language} show that pre-trained models are reasonably well calibrated at predicting whether their own output is factual or not in a few-shot setting.
We leverage their few-shot prompting strategy to elicit numerical confidence of the model in their own predictions.
\citet{tian2023just} find that for instruction tuned models, verbalized probabilities are better-calibrated than conditional probabilities in a setting when models are prompted to output the uncertainty in their predictions.
We refer the reader to \citep{geng2024survey} for a more comprehensive overview of methods to elicit epistemic uncertainty.

\citet{lin2022teaching} introduce the notion of linguistic uncertainty in language models, teaching a pre-trained GPT-3 model to express its uncertainty linguistically along with its prediction on mathematics tasks.
A key difference of our work from theirs is that our uncertainty targets are obtained in a pointwise manner, whereas they assign the targets based on average task performance on questions in the same sub-task.
We also study the effect of the placement of the uncertainty estimate (prefix/suffix) on the quality of the estimates.
\citet{zhou2023navigating} study the effect of prefixed expression of uncertainty on a pre-existing language model’s generation, finding that expression of high certainty hurt the accuracy compared to weaker expressions.
\citet{mielke2022reducing} also work on making models emit linguistic expression of uncertainty, but their process is a two-stage pipeline where a separate calibrator predicts the numerical probability of correctness, and the language model then adds a linguistic marker based on this.
Most similar to our work is \citep{band2024linguistic}, who work on obtaining calibrated long form predictions from LLMs.
Their method of obtaining uncertainty targets relies on a form of self-consistency extended to long form answers, and they also perform an additional reinforcement learning step after their supervised finetuning stage.
They do not study the quality effect of position of the uncertainty estimates relative to the answer (prefix/postfixed).

\section{Setup and Method}
\label{sec:setup}

We investigate the question-answering (Q/A) task where an LLM, specifically Gemini 1.0, incorporates expressions of uncertainty while generating answers.
In a Q/A task, we work with a dataset $\dataset = \{x_i, y_i\}_{i=1}^n$, consisting of $n$ examples where $x$ represents questions and $y$ represents ground-truth answers.
The sets of all questions and answers are denoted by $X = \{x_i\}_{i=1}^n$ and $Y = \{y_i\}_{i=1}^n$, respectively
Given a question $x$, the LLM $M(\cdot)$ produces a prediction $\hat{y} = M(x)$ through autoregressive decoding.
The dataset $\dataset$ is divided into three non-overlapping subsets: few-shot ($\dataset_{fs}$), calibration ($\dataset_{cal}$), and training ($\dataset_{tr}$).
These subsets are used in different phases of finetuning dataset curation -- $\dataset_{fs}$ is used for computing model confidence in the predictions, $\dataset_{cal}$ is used for fitting the isotonic regressor that is then used for post-processing the confidence scores, and $\dataset_{tr}$ is used for finetuning (see \Algref{alg:dataset_curation} for more details).
For each dataset $\dataset$, we reserve a separate held-out subset $\dataset_{te}$ for evaluation purposes.

\subsection{Method} \label{sec:method}
\begin{figure}[t]
    \centering
    \includegraphics[width=1.0\textwidth]{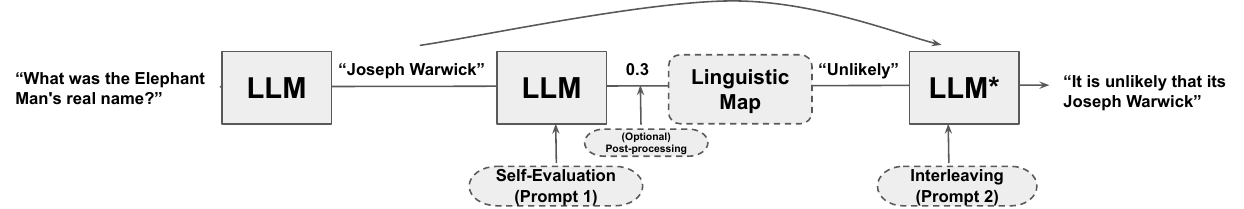}
    \caption{\small \textbf{Finetuning dataset curation process}: Here \texttt{LLM} refers to the language model that we are interested in finetuning. \texttt{LLM}$^*$ referes to an operation that mixes the uncertainty expression with the model prediction -- it can be a prompted language model (interleaved case) or simply an operation which prefixes/post-fixes the answer with the expression of uncertainty. Given a question on the left, the LLM produces a raw prediction and then computes its own confidence on that prediction. The confidence score is converted to a linguistic expression and augmented with the raw prediction. \Promptref{prompt:self_evaluation} and \Promptref{prompt:interleave_uncertainty_answer} are given in the appendix.}
    \label{fig:finetuning_dataset}
\end{figure}

How can we train models to produce accurate expressions of uncertainty?
One approach is few-shot prompting with uncertainty-augmented examples.
However, recent literature suggests that LLMs struggle to generate well-calibrated linguistic expressions of uncertainty through in-context learning \citep{zhou2023navigating}.
Additionally, in-context learning through few-shot prompting incurs extra inference costs for processing prompts with each query, making it less optimal from a latency perspective.
To address these issues, we investigate supervised fine-tuning (SFT) on datasets augmented with uncertainty expressions as an alternative approach.
A critical aspect of curating such datasets is ensuring that the uncertainty markers align with the model’s knowledge about the questions.
Inconsistent uncertainty expressions during fine-tuning can lead to hallucinated expressions during testing.

To ensure linguistic expressions of uncertainty are consistent with the model’s knowledge, we first obtain confidence scores for model-generated samples through self-evaluation, as described by \citet{kadavath2022language}.
Specifically, we use a True/False self-evaluation task, where the model assesses the correctness of its own predictions (see \Promptref{prompt:self_evaluation}).
With the confidence scores in hand, we apply isotonic regression \citep{barlow1972isotonic} using a small calibration set, $\dataset_{cal}$, to achieve nearly perfect calibration of uncertainty estimates.
We then map these calibrated confidence scores to linguistic expressions of uncertainty based on human perception of uncertainty and probabilities \citep{perceptionofprobability}, as detailed in \Tabref{tab:expression_map}.
The final step involves integrating uncertainty expressions with model predictions in the dataset.
We explore three methods for this integration: \textbf{(1) Prefixed:} placing the uncertainty expression before the prediction (\texttt{expression, prediction}), \textbf{(2) Postfixed:} placing the uncertainty expression after the prediction (\texttt{prediction, expression}), and \textbf{(3) Interleaved:} incorporating the uncertainty expression within the prediction (\texttt{expression prediction}) using a prompted language model (see \Promptref{prompt:interleave_uncertainty_answer}).
The overall process of dataset curation is illustrated in \Figref{fig:finetuning_dataset} and \Algref{alg:dataset_curation}.
After curating the dataset $\dataset^{M}$, we fine-tune the model ($M$) to generate linguistic expressions of uncertainty.

\subsection{Evaluation}

\begin{figure}[t]
    \centering
    \includegraphics[width=1.0\textwidth]{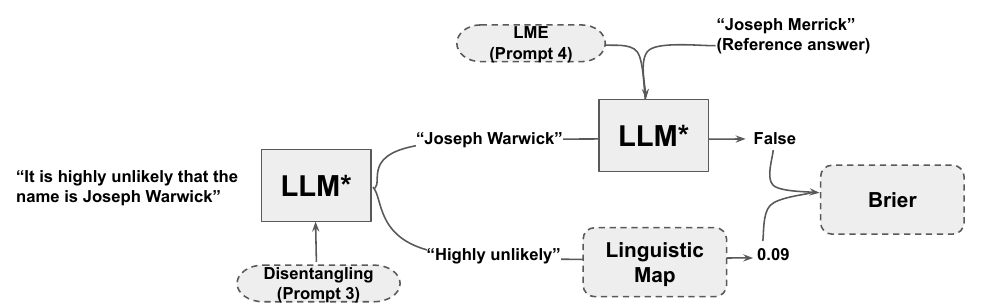}
    \caption{\small \textbf{Evaluation process}: Finetuned LLM produces an answer with the expression of uncertianty on the left that is split by a prompted LLM$^*$ into the raw answer (`Joseph Warwick') and expression of uncertainty (`Highly unlikely') using \Promptref{prompt:deaugmentation}. LLM$^*$ then judges the correctness of the raw answer using the LME~\Promptref{prompt:lme} and uncertainty expression is converted to a float equal to the average of the probability range the uncertainty expression belongs to. Based on the correctness and uncertainty score, the final metric is computed.}
    \label{fig:evaluation}
\end{figure}

\begin{figure}[!t]
    \centering
    \begin{subfigure}[b]{0.23\textwidth}
        \centering
        \includegraphics[width=\textwidth]{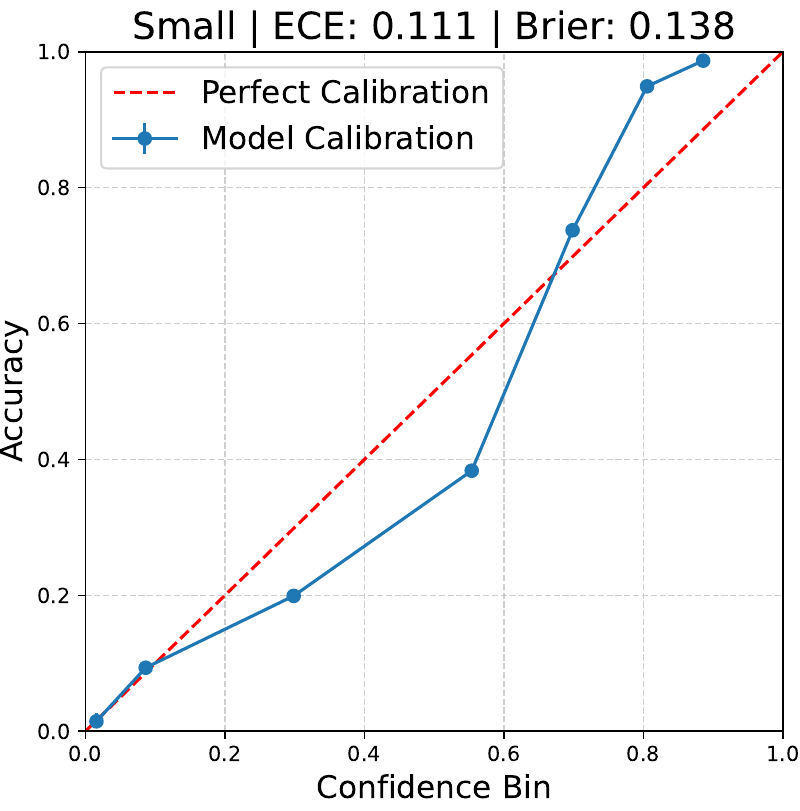}
    \end{subfigure}
    \begin{subfigure}[b]{0.23\textwidth}
        \centering
        \includegraphics[width=\textwidth]{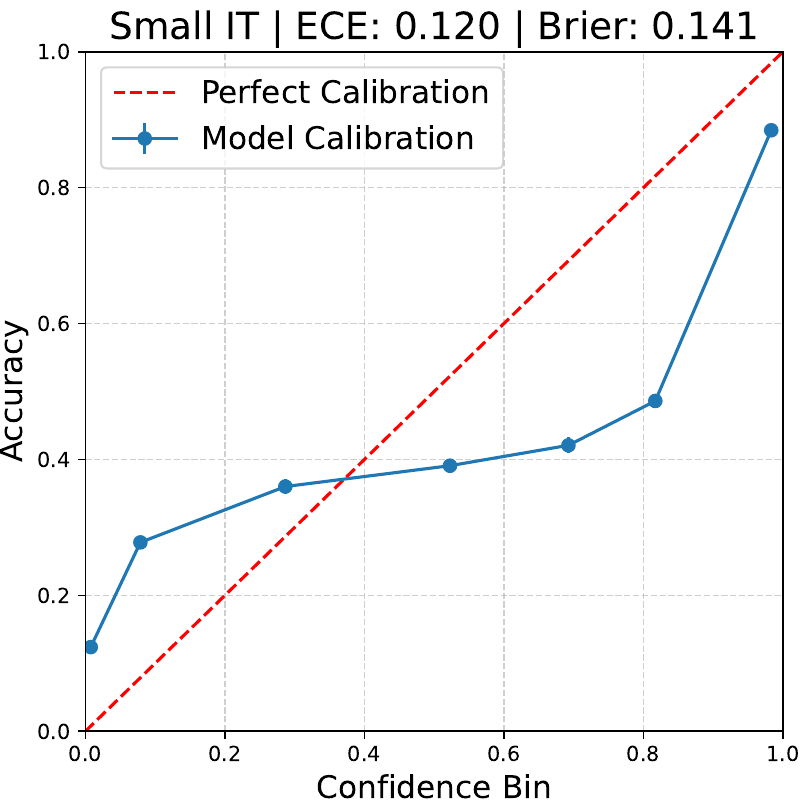}
    \end{subfigure}
    \begin{subfigure}[b]{0.23\textwidth}
        \centering
        \includegraphics[width=\textwidth]{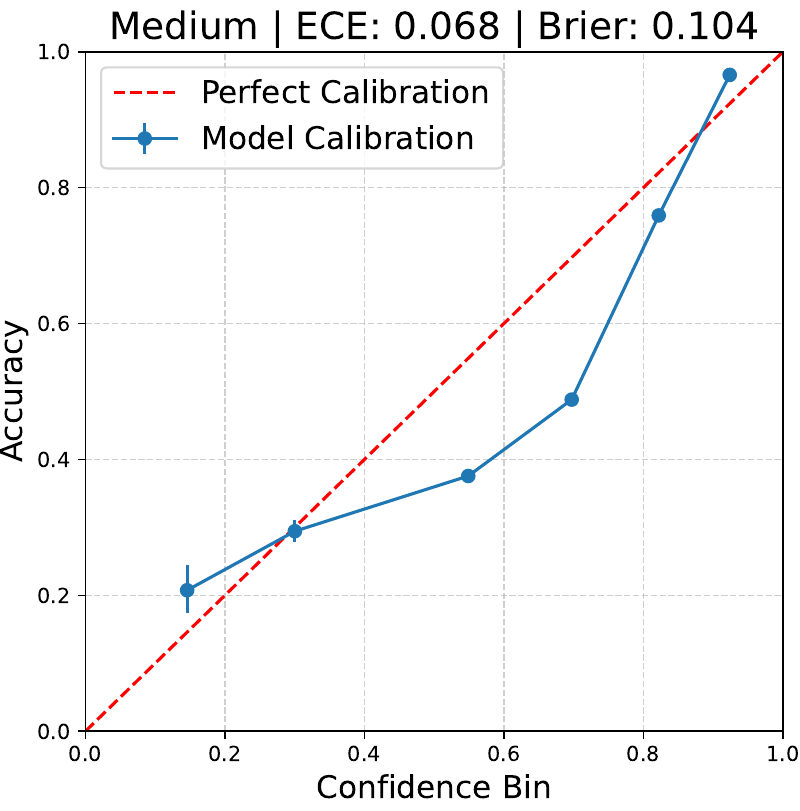}
    \end{subfigure}
    \begin{subfigure}[b]{0.23\textwidth}
        \centering
        \includegraphics[width=\textwidth]{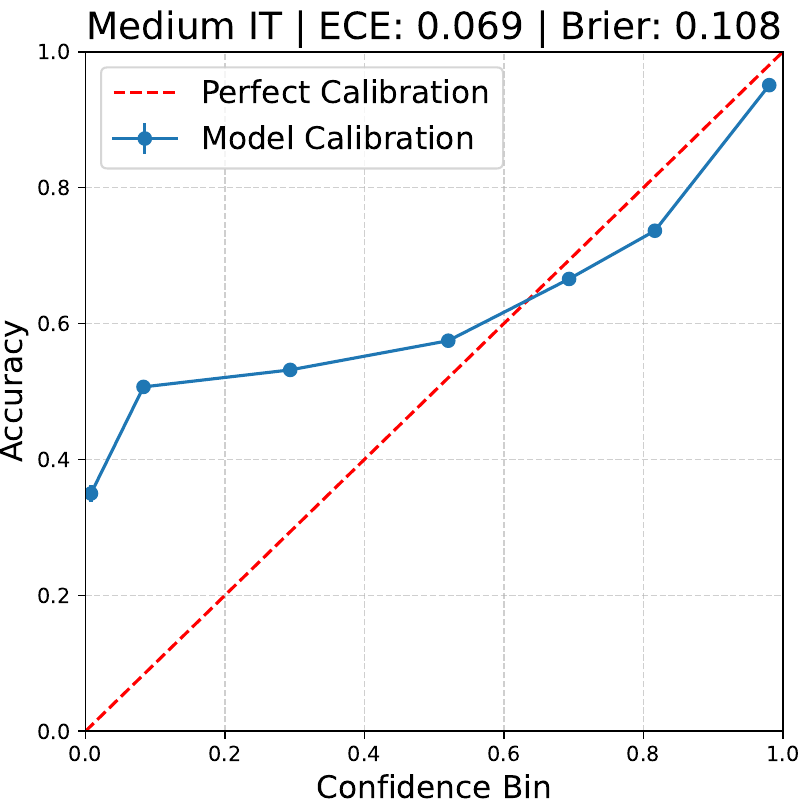}
    \end{subfigure}

    \begin{subfigure}[b]{0.23\textwidth}
        \centering
        \includegraphics[width=\textwidth]{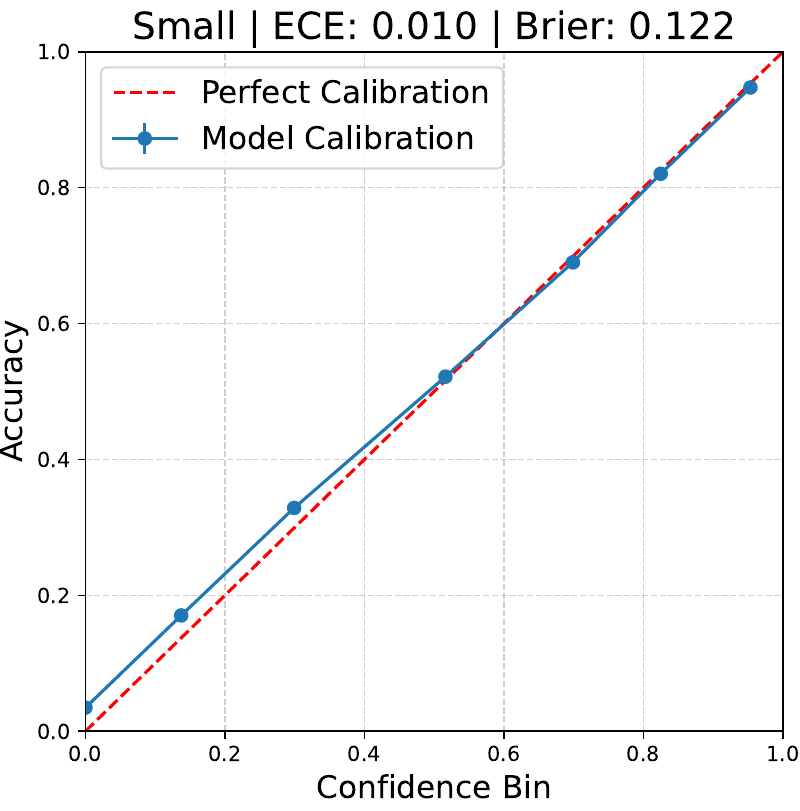}
        \captionsetup{justification=centering}
        \caption{\small \textbf{Small}}
    \end{subfigure}
    \begin{subfigure}[b]{0.23\textwidth}
        \centering
        \includegraphics[width=\textwidth]{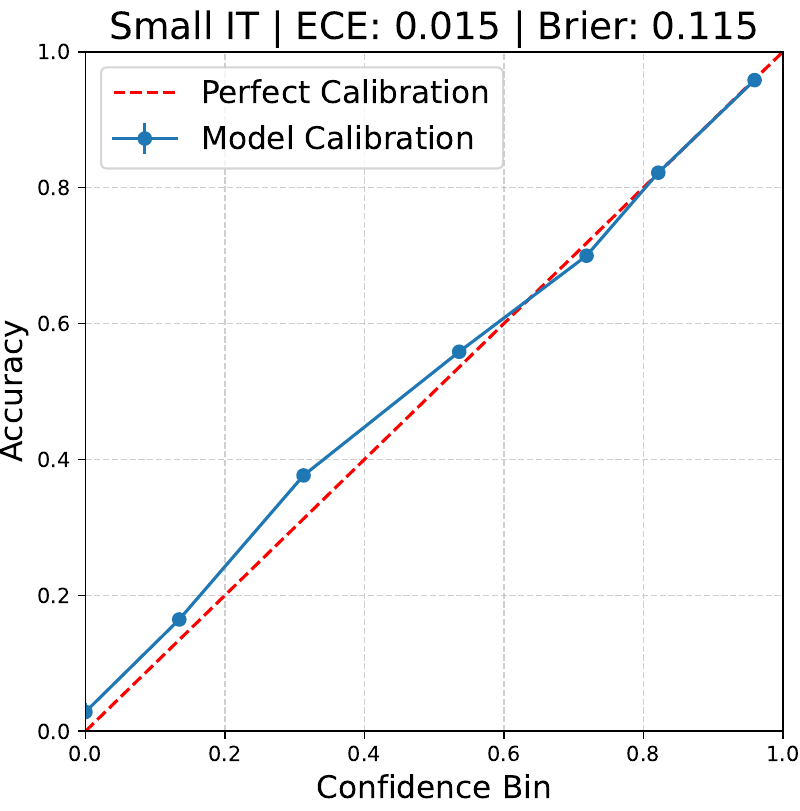}
        \captionsetup{justification=centering}
        \caption{\small \textbf{Small-IT}}
    \end{subfigure}
    \begin{subfigure}[b]{0.23\textwidth}
        \centering
        \includegraphics[width=\textwidth]{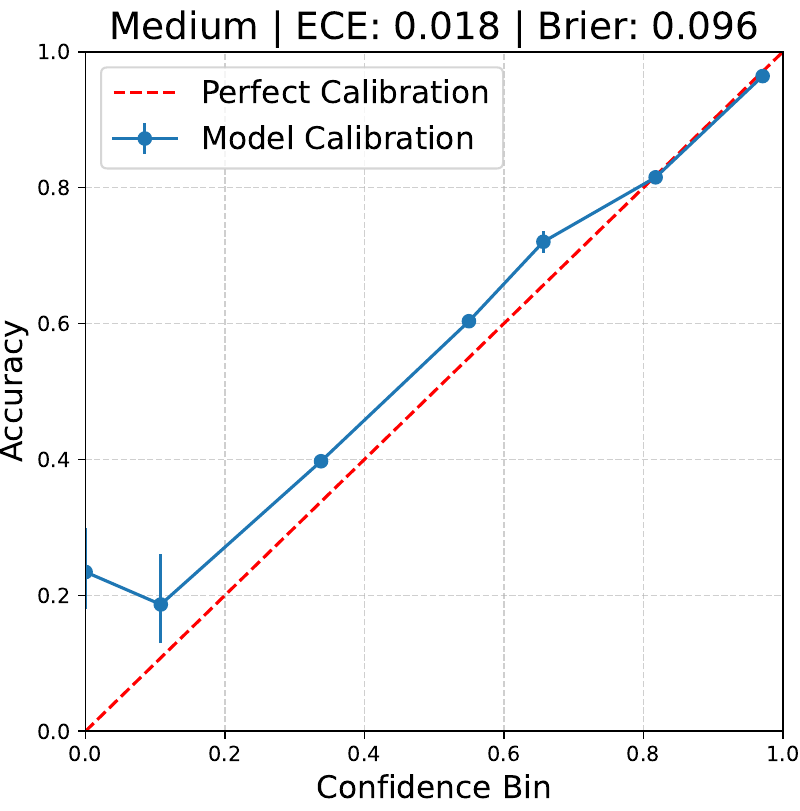}
        \captionsetup{justification=centering}
        \caption{\small \textbf{Medium}}
    \end{subfigure}
    \begin{subfigure}[b]{0.23\textwidth}
        \centering
        \includegraphics[width=\textwidth]{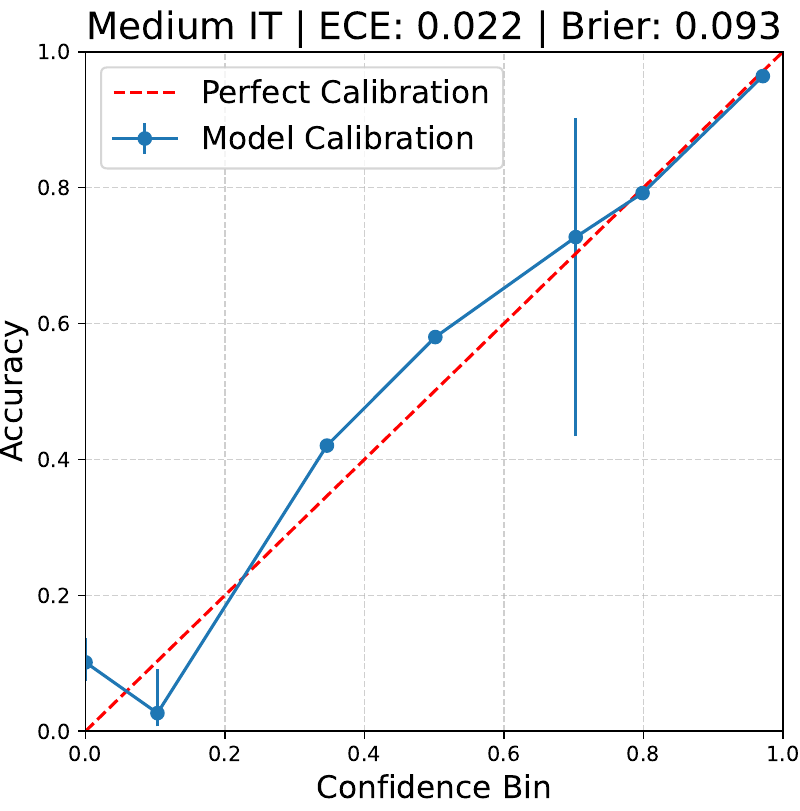}
        \captionsetup{justification=centering}
        \caption{\small \textbf{Medium-IT}}
    \end{subfigure}

    \caption{\textbf{TriviaA Calibration Chart}: The \textbf{top-row} shows raw calibration scores at temperature=1.0 without any post-processing. The \textbf{bottom row} shows post-processed calibration scores with isotonic regression. In each plot, the x-axis is the $p_{model}(true)$ of the generated prediction (shown here as \texttt{Confidence Bin}) and y-axis is probability of that prediction being actually correct (shown here as \texttt{Accuracy}). Expected Calibration Error (ECE) and Brier Score are reported at the top of each plot. The error bars show the variance of accuracy in each bin.}
    \label{fig:trivia_qa_calibration_charts}
\end{figure}

We compute the accuracy (\texttt{Acc}) of a prediction by language model evaluation (\texttt{LME}) whereby a prompted language model compares the prediction with the ground truth answer (see \Promptref{prompt:lme}).
To measure calibration, we plot calibration charts between confidence scores and accuracy (\Figref{fig:trivia_qa_calibration_charts}) where the x-axis is binned according to the probability ranges in the linguistic expressions map~\citep{perceptionofprobability}.
Further, to summarize the calibration error into a single scalar, we track,

\textbf{Expected Calibration Error (ECE):} weighted average of the difference between the confidence assigned to examples in the bin, and accuracy of the predictions in the bin.
$$
ece = \sum_{m=1}^{|B|} \frac{|B_m|}{n} \big|\texttt{Acc}(B_m) - \texttt{C}(B_m)\big|,
$$
where $|B|$ are the total bins,

\textbf{Brier Score:} mean squared error between the confidence and correctness verdicts across all examples
$$
\texttt{brier} = \frac{1}{n} \sum_{i=1}^n \big(\texttt{C}(x_i, \hat{y}_i) - \texttt{LME}(y, \hat{y})\big)^2.
$$
The lower these scores the better. 

Once the models are finetuned to generate linguistic expressions of uncertainty, we test the calibration of their predictions on held-out test sets for each dataset.
For this, we first extract the uncertainty expression from the rest of the prediction using a prompted language model (see \Promptref{prompt:deaugmentation}).
We then convert these uncertainty expressions into probability estimates using the same mapping employed for converting probabilities into linguistic expressions in the previous section.
We measure calibration using Expected Calibration Error (ECE) and Brier Score and plot calibration charts based on these probability estimates.
The complete evaluation procedure is outlined in \Figref{fig:evaluation} and \Algref{alg:eval_uncertainty_predictions}.

\section{Experiments}
\label{sec:experiments}

\subsection{Datasets}
We use standard Q/A datasets -- TriviaQA~\citep{joshi2017triviaqa}, AmbigQA~\citep{min2020ambigqa} and TruthfulQA~\citep{lin2021truthfulqa}.
For \textbf{TriviaQA}, we use the \texttt{wikipedia} version.
From the \texttt{train} split, we take 16 examples for the $\dataset_{fs}$, 2000 examples for the $\dataset_{cal}$, and remaining examples for the $\dataset_{tr}$.
The test set ($\dataset_{te}$) is taken from the \texttt{dev} set.
For \textbf{AmbigQA}, we only take the unambiguous set, as we are focused only on conveying epistemic uncertainty.
From the \texttt{train} split, we take 16 examples for the $\dataset_{fs}$, 1000 examples for the $\dataset_{cal}$, and remaining examples for the $\dataset_{tr}$.
The test set ($\dataset_{te}$) is taken from the \texttt{validation} set.
For \textbf{TruthfulQA}, we split the \texttt{validation} set into a $\dataset_{fs}$ of size 8, $\dataset_{cal}$ of size 100, $\dataset_{te}$ of size 100 and $\dataset_{tr}$ of size \textasciitilde600.
The prediction is deemed correct if the \texttt{LME} (\Promptref{prompt:lme}) judges it to be similar to {\emph any} of the possible ground truth answers.
We evaluate the accuracy of the interleaving (\Promptref{prompt:interleave_uncertainty_answer}), disentangling (\Promptref{prompt:deaugmentation}), and LME (\Promptref{prompt:lme}) prompts on a uniformly sampled subset.
We manually verify the performance of these prompts on small subsets of the data, and find that all of these independent tasks are performed at high accuracy, exceeding 95\%.

\begin{figure}[!t]
    \centering
    \begin{subfigure}[b]{0.28\textwidth}
        \centering
        \includegraphics[width=\textwidth]{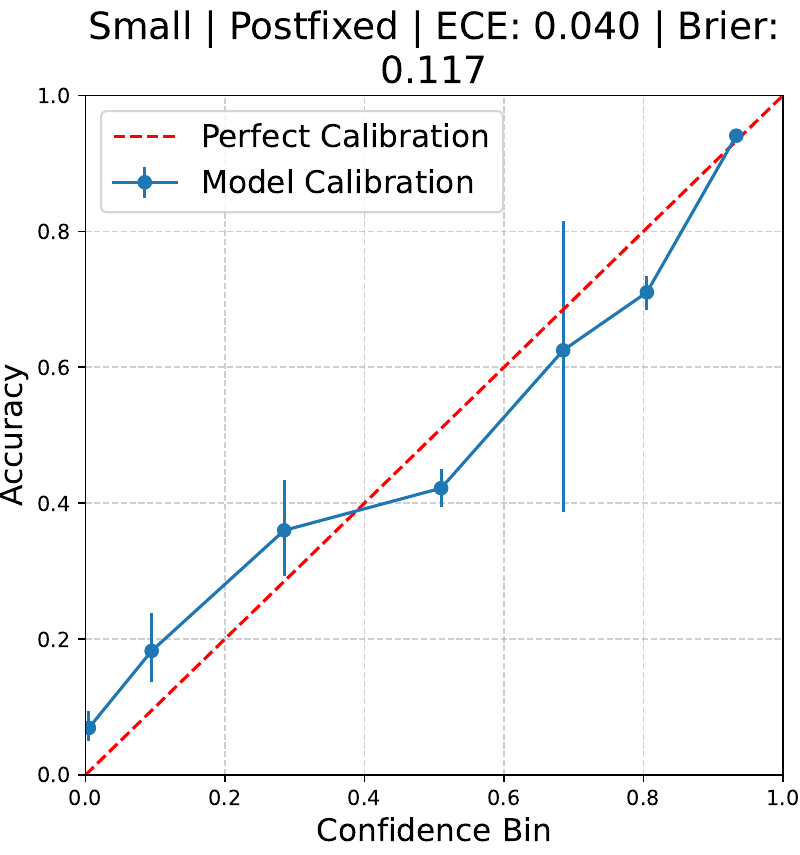}
    \end{subfigure}
    \begin{subfigure}[b]{0.28\textwidth}
        \centering
        \includegraphics[width=\textwidth]{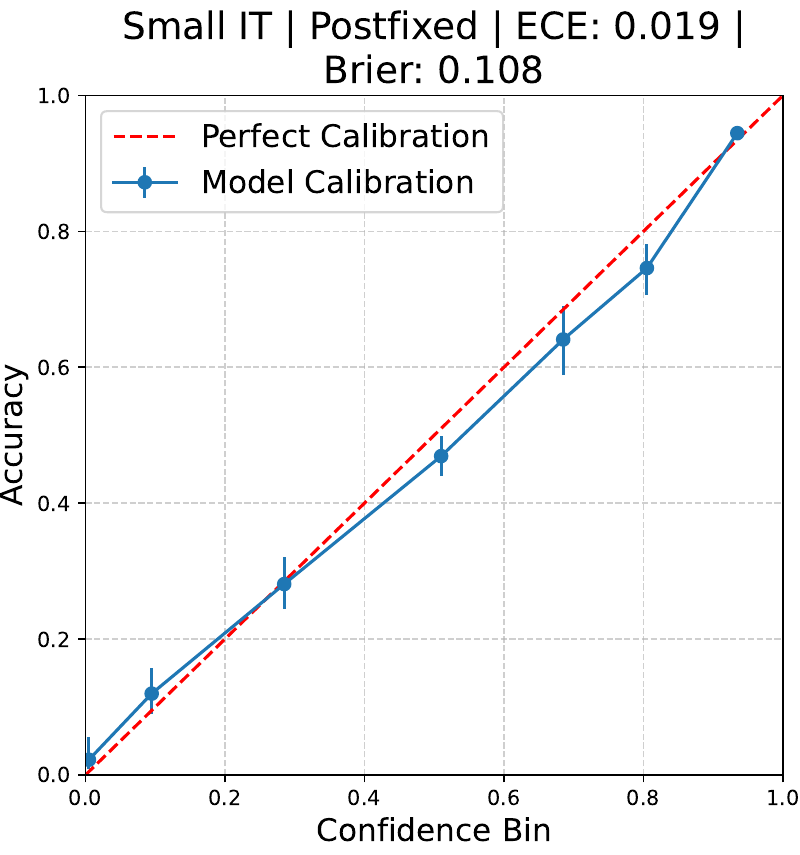}
    \end{subfigure}
    \begin{subfigure}[b]{0.28\textwidth}
        \centering
        \includegraphics[width=\textwidth]{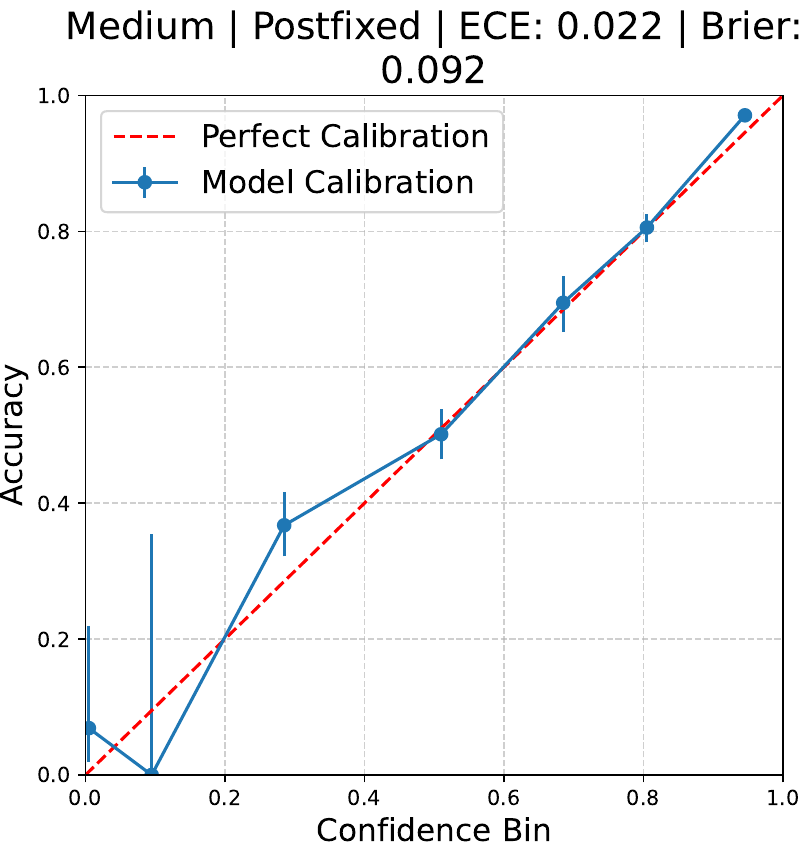}
    \end{subfigure}

    \begin{subfigure}[b]{0.28\textwidth}
        \centering
        \includegraphics[width=\textwidth]{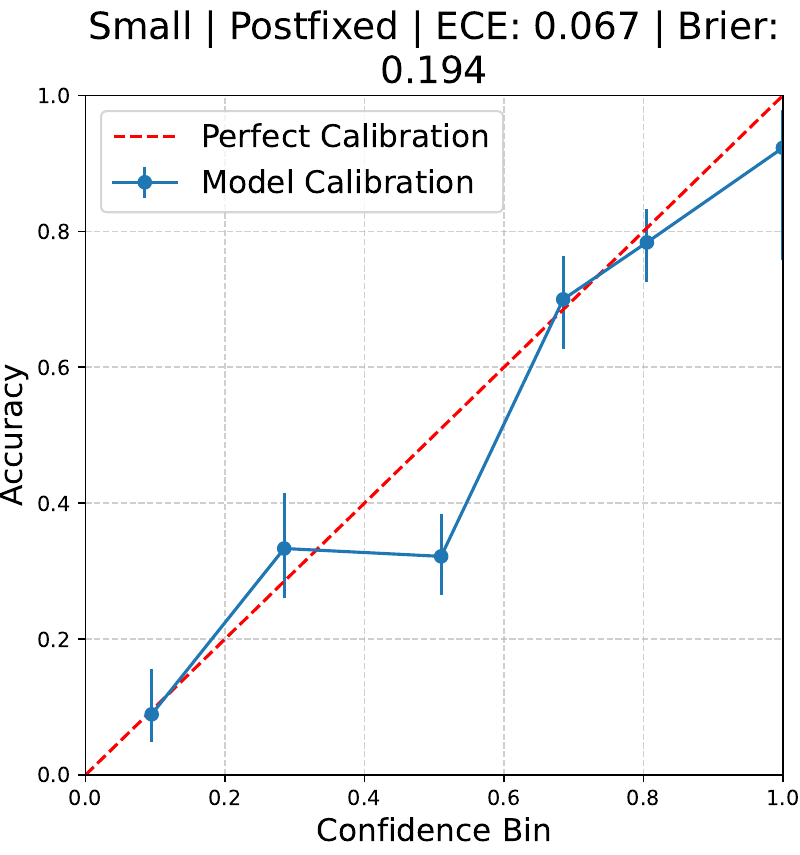}
        \captionsetup{justification=centering}
        \caption{\small \textbf{Small}}
    \end{subfigure}
    \begin{subfigure}[b]{0.28\textwidth}
        \centering
        \includegraphics[width=\textwidth]{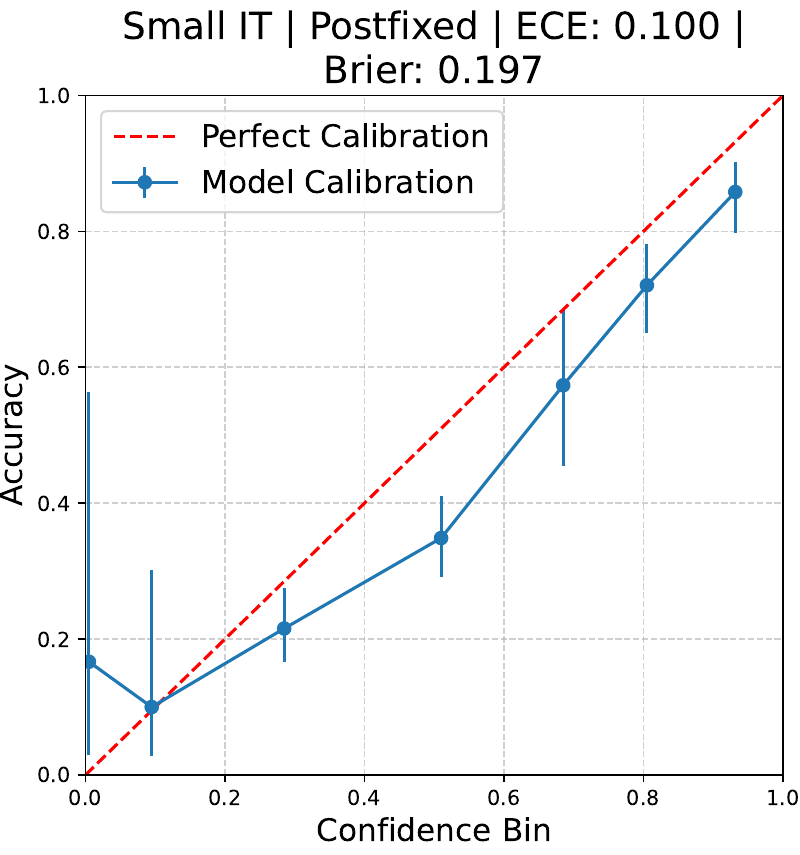}
        \captionsetup{justification=centering}
        \caption{\small \textbf{Small-IT}}
    \end{subfigure}
    \begin{subfigure}[b]{0.28\textwidth}
        \centering
        \includegraphics[width=\textwidth]{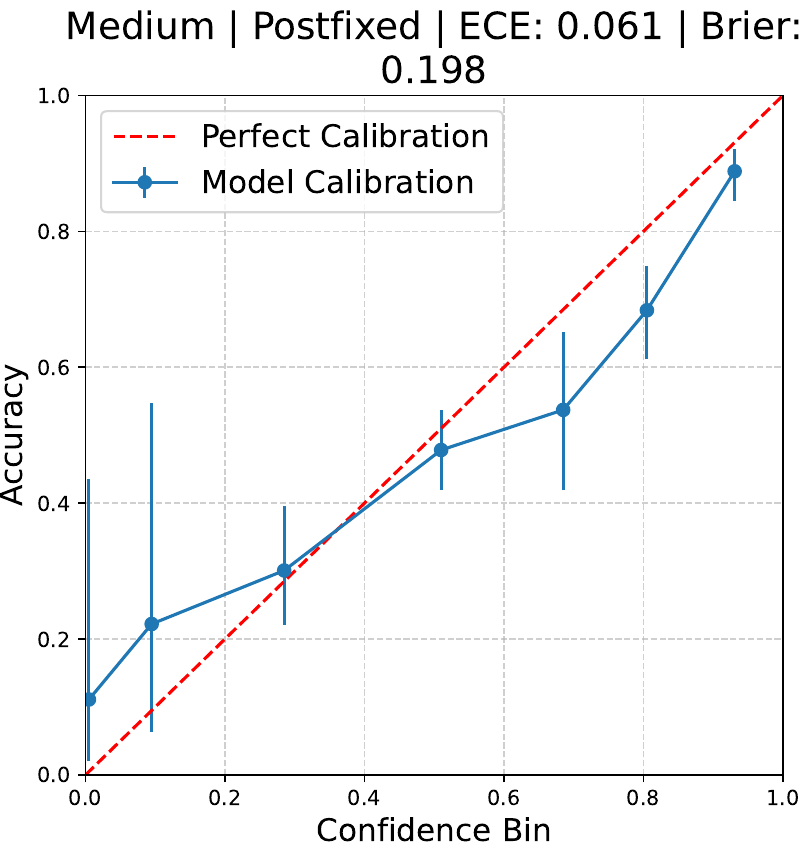}
        \captionsetup{justification=centering}
        \caption{\small \textbf{Medium}}
    \end{subfigure}

    \caption{\textbf{Calibration Charts of Finetuned Models}: Top-row is \textbf{TriviaQA}. Bottom-row is \textbf{AmibQA}. The model generates \textbf{post-fixed} uncertainty expressions. The x-axis is the $p_{model}(true)$ obtained by converting the linguistic expression of uncertainty to a float using \Tabref{tab:expression_map} (shown here as \texttt{Confidence Bin}) and y-axis is probability of that prediction being actually correct (shown here as \texttt{Accuracy}). No post-processing is done on the $p_{model}(true)$. Expected Calibration Error (ECE) and Brier Score are reported at the top of each plot. The error bars show the variance of accuracy in each bin.}
    \label{fig:finetuned_models_calibration_charts}
\end{figure}

\begin{figure}[!t]
    \centering
    \begin{subfigure}[b]{0.28\textwidth}
        \centering
        \includegraphics[width=\textwidth]{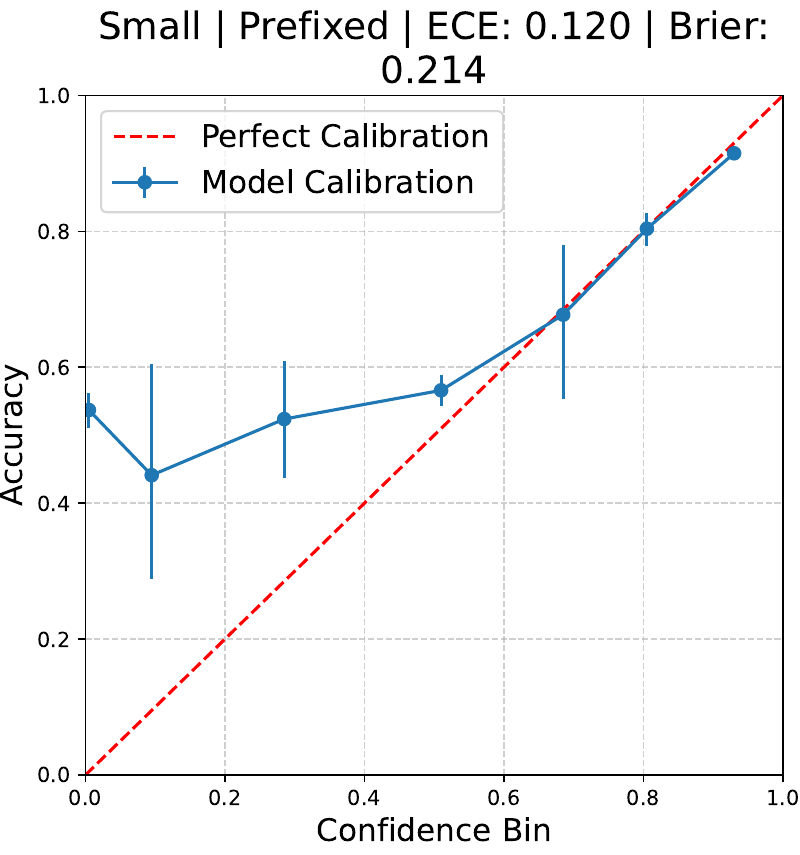}
    \end{subfigure}
    \begin{subfigure}[b]{0.28\textwidth}
        \centering
        \includegraphics[width=\textwidth]{Figures/finetuned/language_pred_prob_calibration_test_set_gemini_small_trivia_qa_postfixed.pdf}
    \end{subfigure}
    \begin{subfigure}[b]{0.28\textwidth}
        \centering
        \includegraphics[width=\textwidth]{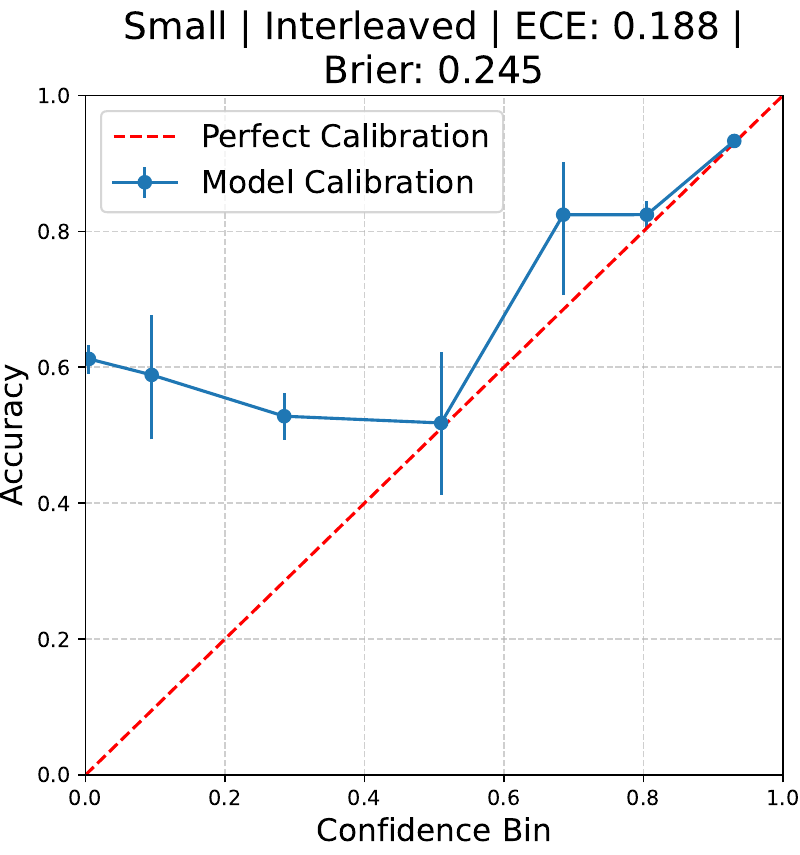}
    \end{subfigure}

    \begin{subfigure}[b]{0.28\textwidth}
        \centering
        \includegraphics[width=\textwidth]{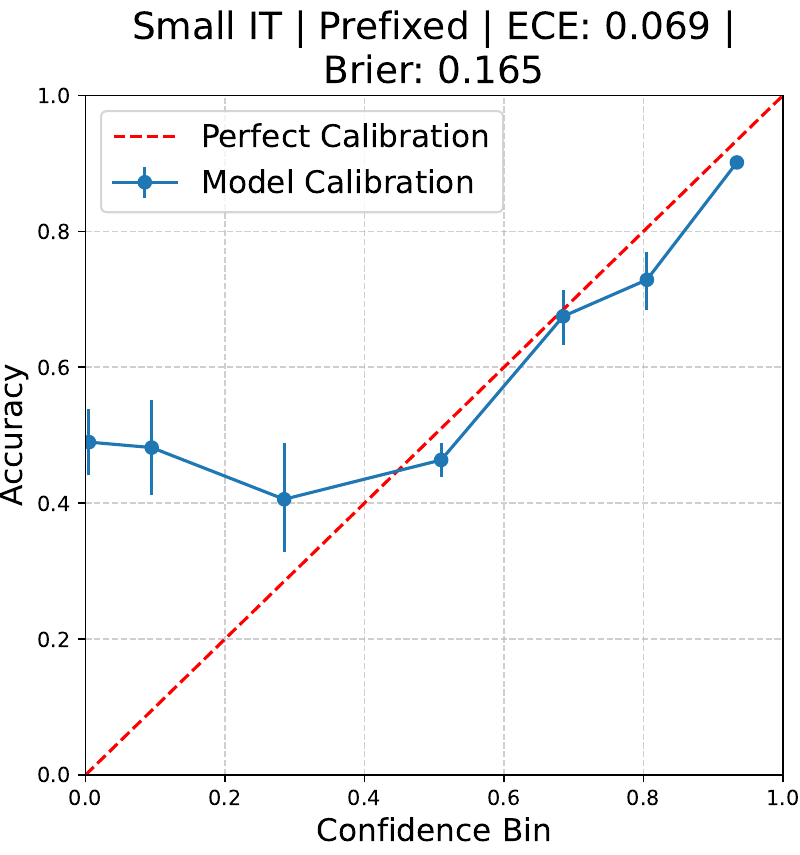}
    \end{subfigure}
    \begin{subfigure}[b]{0.28\textwidth}
        \centering
        \includegraphics[width=\textwidth]{Figures/finetuned/language_pred_prob_calibration_test_set_gemini_small_it_trivia_qa_postfixed.pdf}
    \end{subfigure}
    \begin{subfigure}[b]{0.28\textwidth}
        \centering
        \includegraphics[width=\textwidth]{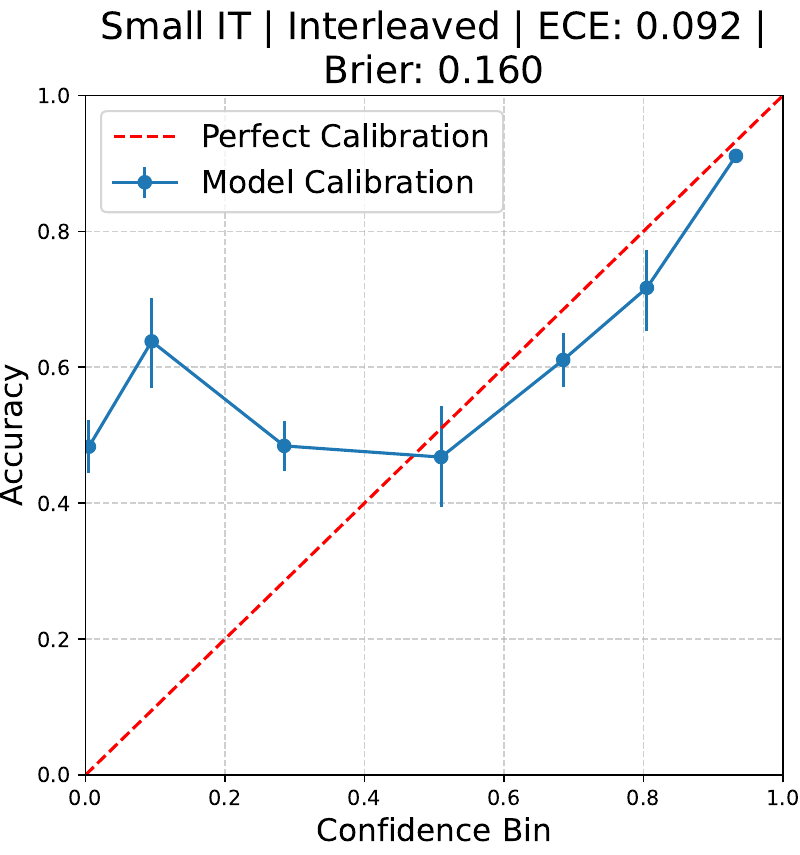}
    \end{subfigure}
    
    \begin{subfigure}[b]{0.28\textwidth}
        \centering
        \includegraphics[width=\textwidth]{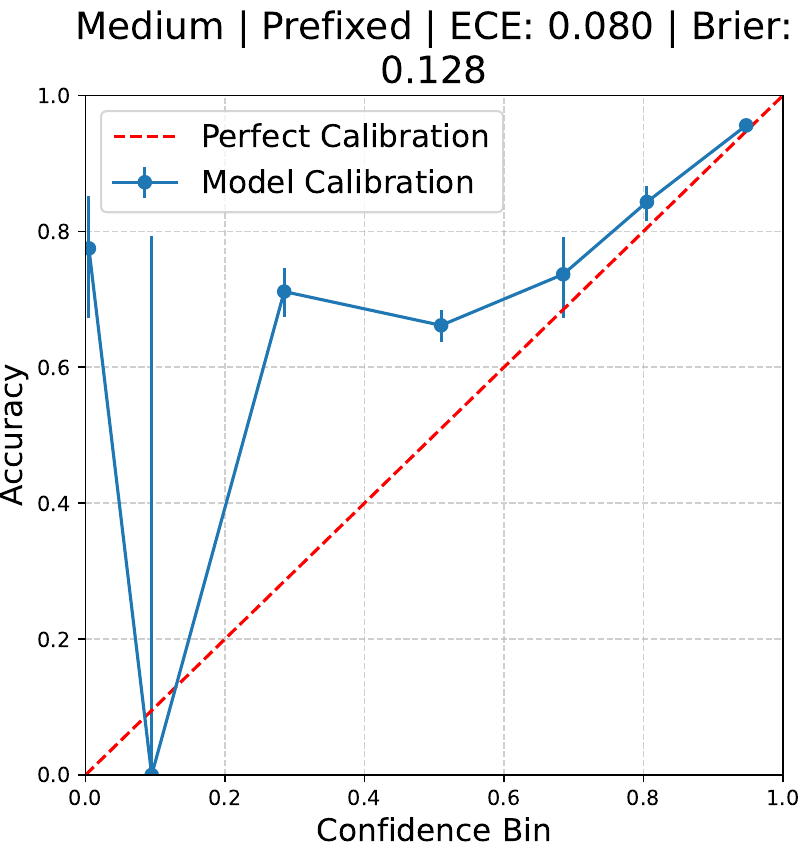}
        \captionsetup{justification=centering}
        \caption{\small \textbf{Prefixed}}
    \end{subfigure}
    \begin{subfigure}[b]{0.28\textwidth}
        \centering
        \includegraphics[width=\textwidth]{Figures/finetuned/language_pred_prob_calibration_test_set_gemini_medium_trivia_qa_postfixed.pdf}
        \captionsetup{justification=centering}
        \caption{\small \textbf{Postfixed}}
    \end{subfigure}
    \begin{subfigure}[b]{0.28\textwidth}
        \centering
        \includegraphics[width=\textwidth]{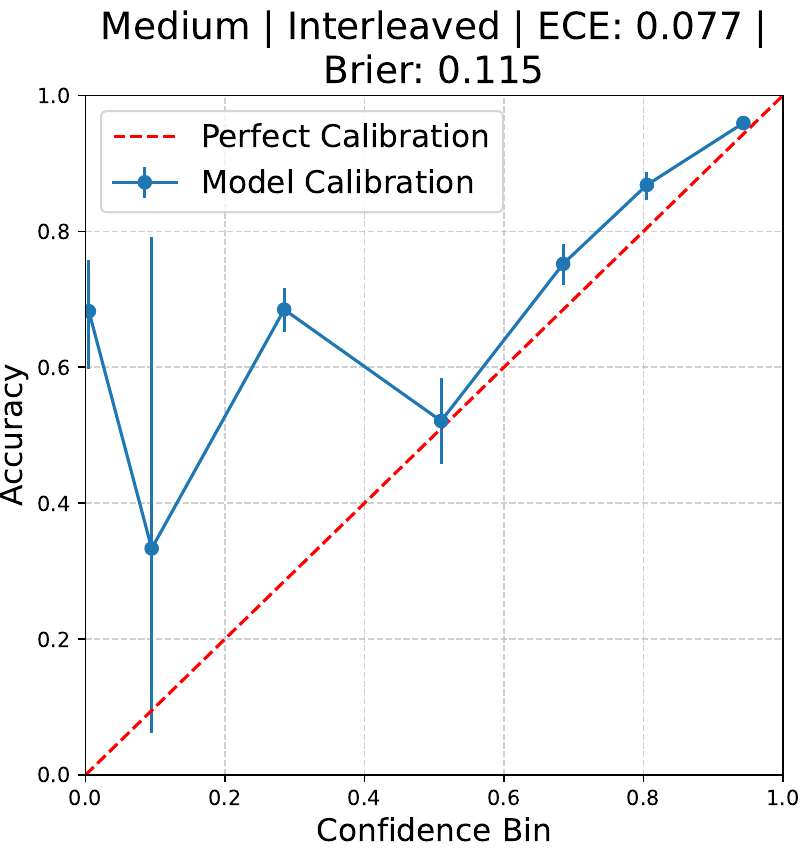}
        \captionsetup{justification=centering}
        \caption{\small \textbf{Interleaved}}
    \end{subfigure}

    \caption{\textbf{TriviaQA uncertainty augmentation method}: Top-row is \textbf{Gemini Small},  middle-row is \textbf{Gemini Small-IT} and the bottom row is \textbf{Gemini Medium}. The models generate \textbf{post-fixed} uncertainty expressions. The x-axis is the $p_{model}(true)$ obtained by converting the linguistic expression of uncertainty to a float using \Tabref{tab:expression_map} (shown here as \texttt{Confidence Bin}) and y-axis is probability of that prediction being actually correct (shown here as \texttt{Accuracy}). No post-processing is done on the $p_{model}(true)$. Expected Calibration Error (ECE) and Brier Score are reported at the top of each plot. The error bars show the variance of accuracy in each bin.}
    \label{fig:trivia_qa_augmentation_methods}
\end{figure}

\subsection{Models}

We conduct experiments with the Gemini 1.0 family of models \citep{team2023gemini}.
Specifically, we use two variants, referred to as \texttt{Small} and \texttt{Medium}, which represent different architecture sizes.
Our study includes both the pre-trained models and those that have been post-trained (aligned).
The post-trained models are denoted with the `\texttt{IT}’ marker.

\subsection{Finetuning Details}

For each question, we independently generate four model samples with a temperature setting of 1.0.
We ensure balance in the curated dataset by setting a maximum number of examples per probability bin (see \Tabref{tab:expression_map}).
Although the curated datasets contain model-generated samples with probability expressions, filtering predictions based on their correctness relative to the ground truth did not yield a significant performance difference.
As a result, we do not apply correctness-based filtering to the datasets in the experiments discussed in this paper.
The statistics for the final curated datasets are provided in \Tabref{tab:datasets_stat}.

We use a batch size of $32$ and train on each dataset for $3$ epochs.
The Adam optimizer \citep{kingma2014adam} with a cosine learning rate schedule is employed, where the learning rate is first linearly warmed up to $5e-7$ and then decayed to $5e-8$.

\subsection{Results}

\textbf{Calibration of Gemini models on the self-evaluation tasks:} \Figref{fig:trivia_qa_calibration_charts} presents the calibration charts for Gemini 1.0 models on the TriviaQA dataset.
Several key observations can be made:
\begin{enumerate}
    \item The Gemini base models exhibit good calibration on the self-evaluation task (\Promptref{prompt:self_evaluation}).
    \item Calibration improves with larger model sizes, and pre-trained models demonstrate better calibration than instruction-tuned models, aligning with findings reported in the literature for other LLMs \citep{kadavath2022language, achiam2023gpt}.
    \item Post-processing confidence scores using isotonic regression leads to significantly improved calibration.
\end{enumerate}
Results for AmbigQA and TruthfulQA are shown in \twoFigref{fig:ambig_qa_calibration_charts}{fig:truthful_qa_calibration_charts}, respectively.
Notably, Gemini models do not achieve good calibration on TruthfulQA using the self-evaluation task.
Although post-processing improves calibration somewhat, it is insufficient for effective uncertainty-augmented dataset curation.
Consequently, we exclude the TruthfulQA dataset from our fine-tuning process.

\paragraph{Calibration of finetuned models:} We now examine the calibration of linguistic expressions of uncertainty in finetuned models on held-out test sets.
\Figref{fig:finetuned_models_calibration_charts} displays the calibration charts for fine-tuned models that produce post-fixed uncertainty expressions.
The figure shows that these fine-tuned models generate well-calibrated linguistic expressions of uncertainty.
This result is consistent across different model sizes and applies equally to both pre-trained and instruction-tuned models.

\paragraph{Comparing different methods of uncertainty augmentation:} Finally, we compare different methods for augmenting model predictions with uncertainty expressions -- prefixed, postfixed, and interleaved -- as previously described.
\Figref{fig:trivia_qa_augmentation_methods} illustrates the performance of these augmentation methods on the TriviaQA dataset, with similar results for AmbigQA shown in \Figref{fig:ambig_qa_augmentation_methods}.
The figure reveals that post-fixed uncertainty expressions, where the uncertainty is added after the main answer, result in the lowest calibration error.
We hypothesize that this approach simplifies fine-tuning because the uncertainty expression does not influence the sampling of the main answer during autoregressive decoding.
In contrast, prefixed or interleaved uncertainty expressions, where the uncertainty is added before or within the answer, can impact the answer’s sampling distribution, as discussed by \citet{zhou2023navigating}, leading to poorer calibration.

\section{Discussion and Conclusion} \label{sec:conclusion}

In this work, we investigated supervised fine-tuning on the model’s own uncertainty as a post-training step to enable models to generate linguistic expressions of uncertainty.
We assessed the calibration of various Gemini 1.0 models and found them to be well-calibrated on the self-evaluation task.
We then used these uncertainty scores to augment model predictions with linguistic expressions of uncertainty.
Our findings show that fine-tuning with these augmented predictions results in models that produce well-calibrated linguistic expressions of uncertainty on held-out test sets.
This fine-tuning approach can be employed as an independent post-training step between supervised fine-tuning (SFT) and reinforcement learning from human feedback (RLHF).
Alternatively, uncertainty-augmented datasets can be integrated into the SFT process with appropriate system instructions that can guide models to express uncertainty.

Models capable of generating well-calibrated uncertainty expressions enable users to make informed inferences about the model’s predictions.
With linguistic expressions of uncertainty, users can reliably decide when to trust the model’s predictions and when to seek additional information.
This allows users greater control over how to utilize the model’s outputs, unlike methods that rely on uncertainty estimates to determine when to abstain from giving a response.
By abstaining, these methods may deprive users of potentially valuable information, even when the model’s responses are uncertain.
Therefore, the ability to produce well-calibrated expressions of uncertainty should be considered a key objective in the development of user facing foundational models.

\section{Acknowledgements} 

We extend our gratitude to Greg Wayne, Reed Roberts, and Mehdi Bennani for their early discussions on the project.
We also thank Taylan Cemgil and Jacob Eisenstein for their detailed feedback on the draft.

\newpage
\bibliography{main}
\newpage
\appendix
\onecolumn
\section{Appendix}

\begin{algorithm}[h]
\caption{Dataset curation with linguistic expressions of uncertainty}\label{alg:dataset_curation}
\begin{algorithmic}[1]

\State \textbf{Input:} Main Model ($M$), Augmentation Model ($F$), Confidence Routine (\texttt{Confidence}), Confidence Score to Linguistic Expression Map (\texttt{LinguisticMap}), Isotonic Regression Routine (\texttt{IsoReg}), Dataset ($\dataset = \{\dataset_{tr},\dataset_{cal},\dataset_{fs}\})$
\State \textbf{Output:} A curated dataset ($\dataset_{tr}^M$)

\State $\hat{Y} = M(X)$ \algorithmiccomment{Compute model predictions.}
\State $C_{cal} = \texttt{Confidence}(X_{cal}, \hat{Y}_{cal}, \dataset_{fs})$ \algorithmiccomment{Compute confidence scores on calibration set.}
\State $C_{tr} = \texttt{Confidence}(X_{tr}, \hat{Y}_{tr}, \dataset_{fs})$ \algorithmiccomment{Compute confidence scores on train set.}
\State $R = \texttt{IsoReg}(Y_{cal}, \hat{Y}_{cal}, C_{cal})$ \algorithmiccomment{Fit an isotonic regressor on the calibration set.}
\State $C_{tr}^* = R(C_{tr})$ \algorithmiccomment{Post-process confidence scores using the regressor.}
\State $E_{tr} = \texttt{LinguisticMap}(C_{tr}^*)$ \algorithmiccomment{Convert confidence scores to linguistic expressions of uncertainty.}
\State $\dataset_{tr}^M = F(X_{tr}, \hat{Y}_{tr}, E_{tr})$ \algorithmiccomment{Augment model predictions with linguistic expressions of uncertainty.}

\State \textbf{Return} $\dataset_{tr}^M$

\end{algorithmic}
\end{algorithm}

\begin{algorithm}[h]
\caption{Evaluation of uncertainty augmented predictions}\label{alg:eval_uncertainty_predictions}
\begin{algorithmic}[1]

\State \textbf{Input:} Dataset containing uncertainty augmented predictions $\dataset_{te} = (X, \Tilde{Y})$, Deaugmentation Model (\texttt{DeAug}), a map to convert uncertainty statements to floats (\texttt{InvMap}), Routine to compute ECE (\texttt{ECE}), Routine to compute Brier Score (\texttt{Brier Score})
\State \textbf{Output:} $ece$, $brier$

\State $\hat{Y}, E = F(\Tilde{Y})$ \algorithmiccomment{Strip the uncertainty statements and main answers.}
\State $C = \texttt{InvMap}(E)$ \algorithmiccomment{Map uncertainty statements to floats.}
\State $ece = \texttt{ECE}(Y, \hat{Y}, C)$
\State $brier = \texttt{Brier Score}(Y, \hat{Y}, C)$

\State \textbf{Return} $ece$, $brier$

\end{algorithmic}
\end{algorithm}

\begin{table}[h]
\centering
\begin{small}
\begin{tabular}{l l}
\specialrule{1.5pt}{0pt}{0pt}
\texttt{\textbf{Probability Range}} & \texttt{\textbf{Linguistic Expressions}} \\
\specialrule{1.5pt}{0pt}{0pt}
\texttt{[1.0-1.0]} & \texttt{(`certain')} \\ 
\texttt{[0.87-0.99]} & \texttt{(`almost certain', `highly likely')} \\ 
\texttt{[0.75-0.86]} & \texttt{(`very good chance', `likely')} \\ 
\texttt{[0.63-0.74]} & \texttt{(`i believe', `probably', `probable')} \\ 
\texttt{[0.40-0.62]} & \texttt{(`chances are about even')} \\ 
\texttt{[0.18-0.39]} & \texttt{(`i doubt', `unlikely')} \\ 
\texttt{[0.02-0.17]} & \texttt{(`little chance',`chances are slight',`improbable',`highly unlikely')} \\ 
\texttt{[0.00-0.01]} & \texttt{(`impossible')} \\ 
\specialrule{1.5pt}{0pt}{0pt}
\end{tabular}
\end{small}
\caption{\textbf{Probability to Linguistic Expression Map}}
\label{tab:expression_map}
\end{table}

\begin{lstlisting}[style=mystyle, language=bash, caption={\textbf{Few-shot prompt for self-evaluation.}}, label={prompt:self_evaluation}]
You are a highly knowledgeable assistant who is always factual and careful. For a {QUESTION} check if the {PROPOSED_ANSWER} is True
or False.

{QUESTION}: Who was the first president of the US?
{PROPOSED_ANSWER}: George Washington
Is the Proposed Answer:
a) True
b) False
The proposed Answer is: True

...

{QUESTION}: <query>
{PROPOSED_ANSWER}: <model prediction>
Is the Proposed Answer:
a) True
b) False
The proposed Answer is: 
\end{lstlisting}

\begin{lstlisting}[style=mystyle, language=bash, caption={\textbf{Prompt to interleave uncertainty expressions with answers.}}, label={prompt:interleave_uncertainty_answer}]
You are a powerful agent that understands probabilities and language. Given a {QUESTION}, {ANSWER} and {PROBABILITY_EXPRESSION}, you
modify the {ANSWER} using the appropriate probability expression. Do not modify the core content of the {ANSWER}, and the only
change made is to include the {PROBABILITY_EXPRESSION} to the {ANSWER} to produce the {MODIFIED_ANSWER}.

Here are a few examples:

Example 1:
{QUESTION}: Which English author, who died in 1998, published Katie Hannigan in 1950 and The Bill Bailey Trilogy between 1986 and
1997?
{ANSWER}: richard adamson
{PROBABILITY_EXPRESSION}: little chance
{MODIFIED_ANSWER}: There is little chance that the author is Richard Adamson.

Example 2:
{QUESTION}: What chemical element has the symbol S?
{ANSWER}: sulphur
{PROBABILITY_EXPRESSION}: very good chance
{MODIFIED_ANSWER}: There is a very good chance that it is sulphur.

Example 3:
{QUESTION}: What is secreted by the lachrymal glands?
{ANSWER}: tears
{PROBABILITY_EXPRESSION}: certain
{MODIFIED_ANSWER}: It is certainly tears.

Example 4:
{QUESTION}: How many days is the gestation period of a domestic dog (5 days either way)?
{ANSWER}: 58-63
{PROBABILITY_EXPRESSION}: chances are about even
{MODIFIED_ANSWER}: Chances are about even that it is 58-63 days.

Example 5:
{QUESTION}: Which record label signed the Rolling Stones in 1991?
{ANSWER}: umg
{PROBABILITY_EXPRESSION}: i doubt
{MODIFIED_ANSWER}: I doubt that it is umg.

Example 6:
{QUESTION}: What is the capital of India?
{ANSWER}: Delhi.
{PROBABILITY_EXPRESSION}: impossible
{MODIFIED_ANSWER}: It is impossible that it is Delhi.

Example 7:

{QUESTION}: {{THE_QUESTION}}
{ANSWER}: {{THE_ANSWER}}
{PROBABILITY_EXPRESSION}: {{THE_PROBABILITY_EXPRESSION}}
{MODIFIED_ANSWER}: 

\end{lstlisting}

\begin{lstlisting}[style=mystyle, language=bash, caption={\textbf{Prompt to disentangle uncertainty statements from answers.}}, label={prompt:deaugmentation}]
You are presented with a statement, which may contain a notion of uncertainty expressed linguistically within it. Your task is to
extract the {UNCERTAINTY_PHRASE} separately, and remove the uncertainty component from the answer.

Here is the list of valid uncertainty expressions.

[certain, almost certain, highly likely, very good chance, likely, i believe, probably, probable, chances are about even,
i doubt, unlikely, little chance, chances are slight, improbable, highly unlikely, impossible].

List the Answer removed of uncertainty in the {ANSWER_WITHOUT_UNCERTAINTY} field. List the uncertainty expression used in the
Uncertainty field.

{ANSWER}: X was certainly not born in 1985.
{ANSWER_WITHOUT_UNCERTAINTY}: X was not born in 1985.
{UNCERTAINTY_PHRASE}: certainly

{ANSWER}: The capital of France almost certainly might be paris.
{ANSWER_WITHOUT_UNCERTAINTY}: The capital of France is Paris.
{UNCERTAINTY_PHRASE}: almost certainly

{ANSWER}: There is little chance but the fact is correct.
{ANSWER_WITHOUT_UNCERTAINTY}: The fact is correct.
{UNCERTAINTY_PHRASE}: little chance

{ANSWER}: It is about even that the coin will be heads.
{ANSWER_WITHOUT_UNCERTAINTY}: The coin will be heads.
{UNCERTAINTY_PHRASE}: about even

{ANSWER}: It is impossible that the Sun rises in the West.
{ANSWER_WITHOUT_UNCERTAINTY}: The Sun rises in the West.
{UNCERTAINTY_PHRASE}: impossible

{ANSWER}: It is highly unlikely that the coin will be heads.
{ANSWER_WITHOUT_UNCERTAINTY}: The coin will be heads.
{UNCERTAINTY_PHRASE}: highly unlikely

{ANSWER}: There is a very good chance that tomato might not be a vegetable.
{ANSWER_WITHOUT_UNCERTAINTY}: Tomato is not a vegetable.
{UNCERTAINTY_PHRASE}: very good chance

Here is the {ANSWER} which needs to be separated into {ANSWER_WITHOUT_UNCERTAINTY}, and {UNCERTAINTY_PHRASE}.

{ANSWER}:
\end{lstlisting}

\begin{lstlisting}[style=mystyle, language=bash, caption={\textbf{Language model evaluation.}}, label={prompt:lme}]
Your task is to determine if two answers to a question are semantically equivalent. Two answers are semantically equivalent if their
answer to the question is the same, even if they are rephrases of each other.

Even if one contains more information than the other, as long as their answer to the question is the same, the answers are considered
semantically equivalent.

For example, for the question `Tell me a number`, the answers `five` and `6` are not semantically equivalent as they are different
numbers. However, the answers `5` and `A number is five` or `five` are semantically equivalent, since they convey the same thing.

For a question `Tell me the capital of France`, `Its Venice` and `Venice` are semantically equivalent, as they give the same answer.
The Answers `Paris` and `Venice` are not semantically equivalent. For questions asking factual information, if one answer responds
with information disagreeing with the other, they will not be semantically equivalent.

Another example of not semantically equivalent answers would be `Miles` and `Kilometers` for the question `Tell me an unit of
distance`, as these two units are not the same.

Here is the question, and Answer A and Answer B to be compared.

Respond with `YES` if they are semantically equivalent, `NO` otherwise.

Question: {THE_QUESTION}
Answer A: {GOLD_ANSWER}
Answer B: {PRED_ANSWER}
Semantically equivalent:
\end{lstlisting}

\begin{table}[h]
\centering
\begin{small}
\begin{tabular}{l l c c c}
\specialrule{1.5pt}{0pt}{0pt}
\textbf{Model} & \textbf{Dataset} & \textbf{Max Examples Per Bin} & \textbf{Train Set Size} & \textbf{Evaluation Set size} \\
\specialrule{1.5pt}{0pt}{0pt} 
\multirow{3}{*}{Small} & TriviaQA & 2000 & 12556 & 7993 \\
& AmbigQA & 1000 & 8935 & 898 \\
& TruthfulQA & - & - & - \\
\hline
\multirow{3}{*}{Small - IT} & TriviaQA & 2000 & 11571 & 7993 \\
& AmbigQA & 1000 & 8487 & 898 \\
& TruthfulQA & 200 & 581 & 100 \\
\hline
\multirow{3}{*}{Medium} & TriviaQA & 2000 & 9283 & 7993 \\
& AmbigQA & 1000 & 4800 & 898 \\
& TruthfulQA & 200 & 440 & 100 \\
\hline
\multirow{3}{*}{Medium - IT} & TriviaQA & 2000 & 7590 & 7993 \\
& AmbigQA & 1000 & 3969 & 898 \\
& TruthfulQA & 200 & 518 & 100 \\
\specialrule{1.5pt}{0pt}{0pt}
\end{tabular}
\end{small}
\caption{\textbf{Statistics of the datasets used for finetuning $\dataset_{tr}^M$}}
\label{tab:datasets_stat}
\end{table}

\begin{figure}[htp]
    \centering
    \begin{subfigure}[b]{0.23\textwidth}
        \centering
        \includegraphics[width=\textwidth]{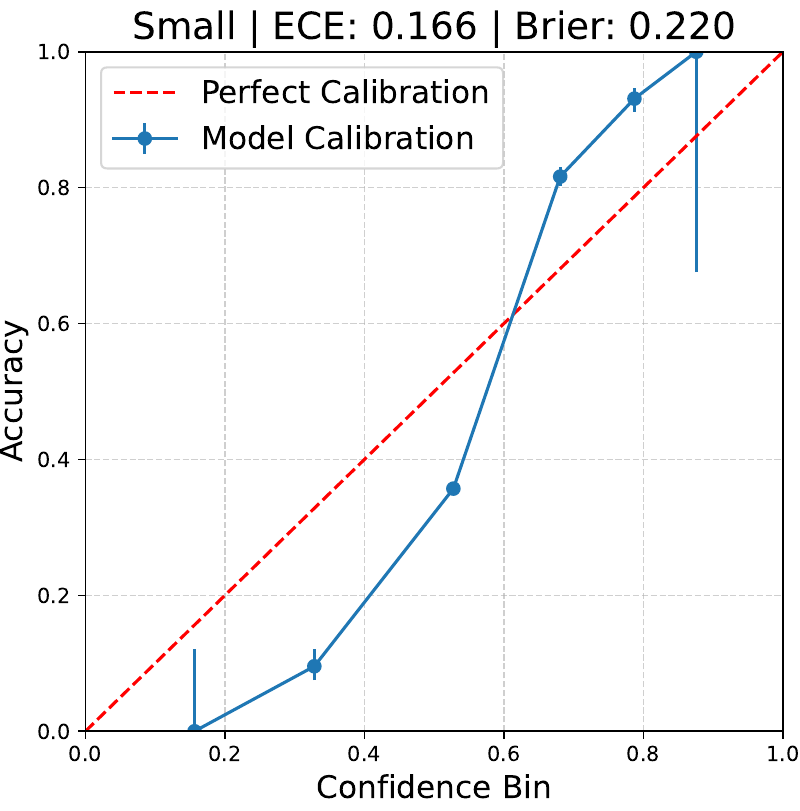}
    \end{subfigure}
    \begin{subfigure}[b]{0.23\textwidth}
        \centering
        \includegraphics[width=\textwidth]{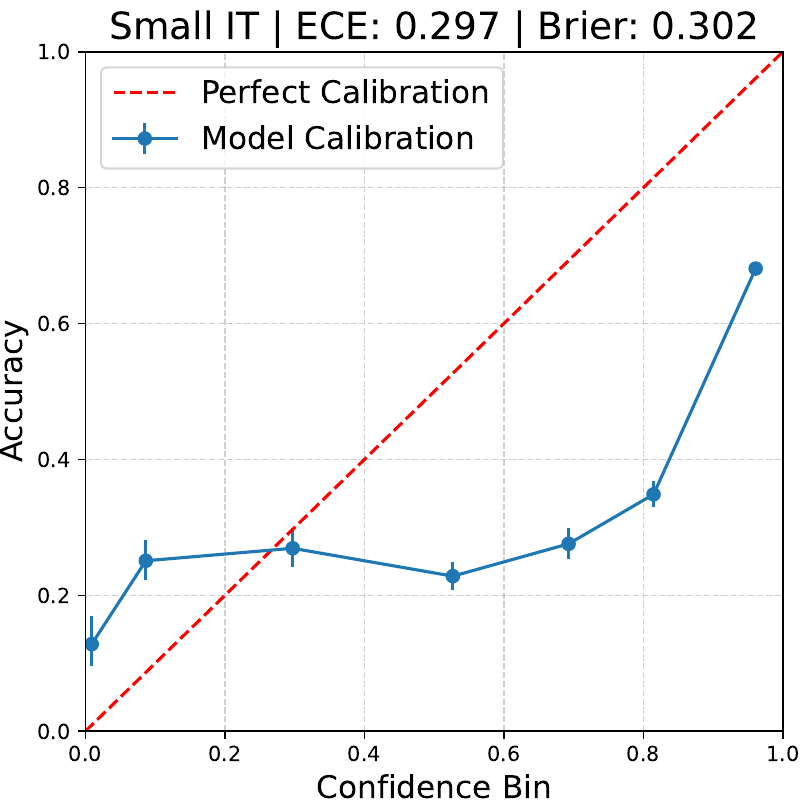}
    \end{subfigure}
    \begin{subfigure}[b]{0.23\textwidth}
        \centering
        \includegraphics[width=\textwidth]{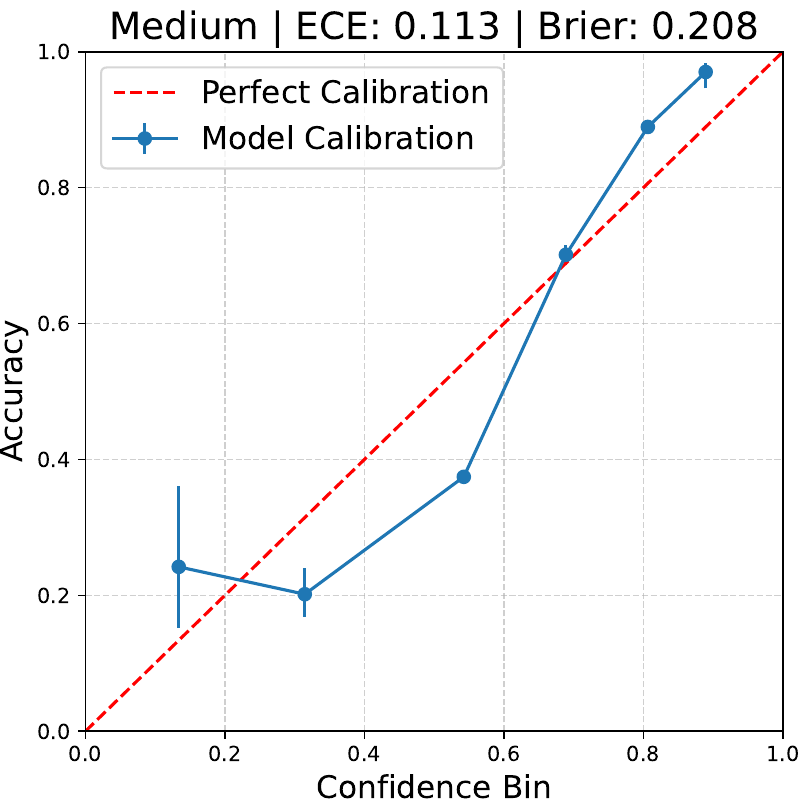}
    \end{subfigure}
    \begin{subfigure}[b]{0.23\textwidth}
        \centering
        \includegraphics[width=\textwidth]{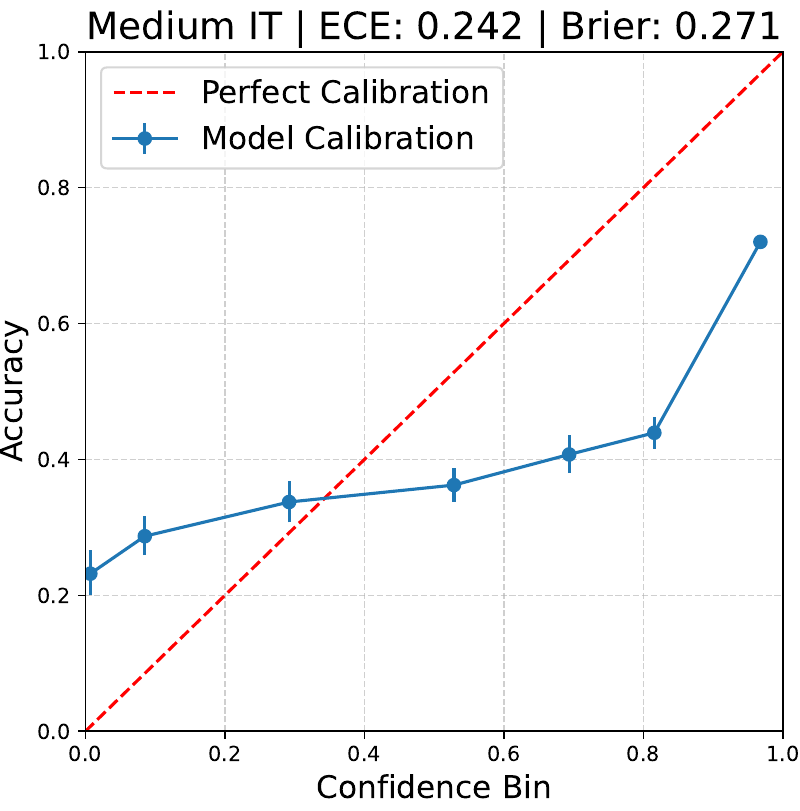}
    \end{subfigure}

    \begin{subfigure}[b]{0.23\textwidth}
        \centering
        \includegraphics[width=\textwidth]{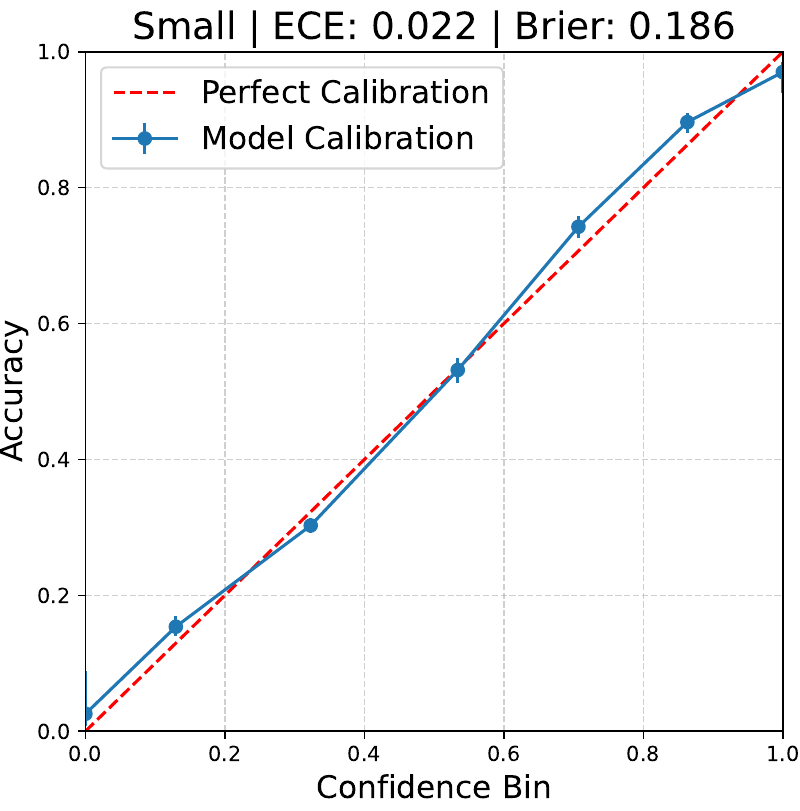}
        \captionsetup{justification=centering}
        \caption{\small \textbf{Small}}
    \end{subfigure}
    \begin{subfigure}[b]{0.23\textwidth}
        \centering
        \includegraphics[width=\textwidth]{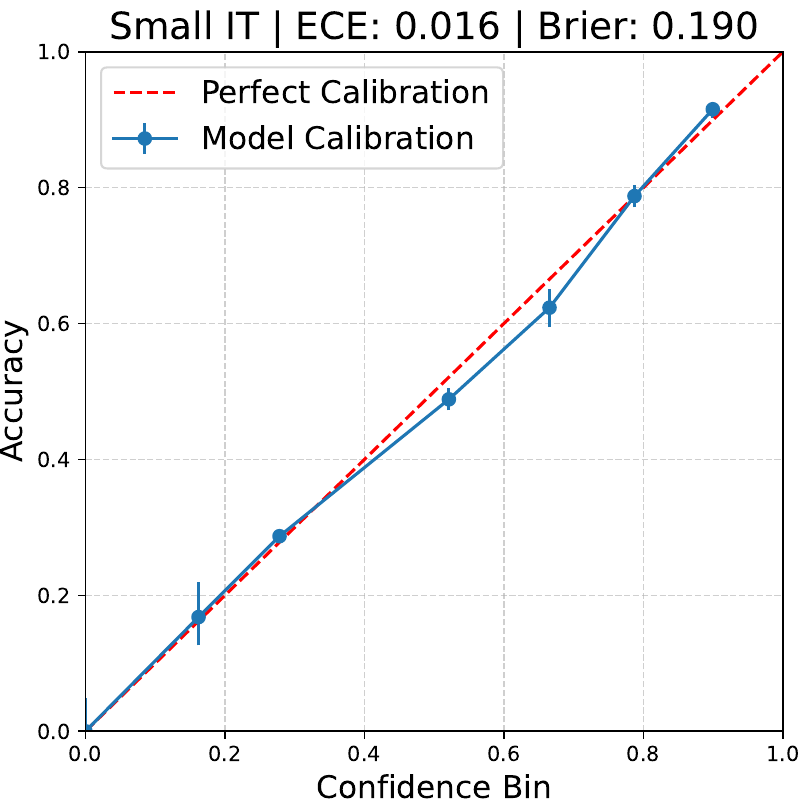}
        \captionsetup{justification=centering}
        \caption{\small \textbf{Small-IT}}
    \end{subfigure}
    \begin{subfigure}[b]{0.23\textwidth}
        \centering
        \includegraphics[width=\textwidth]{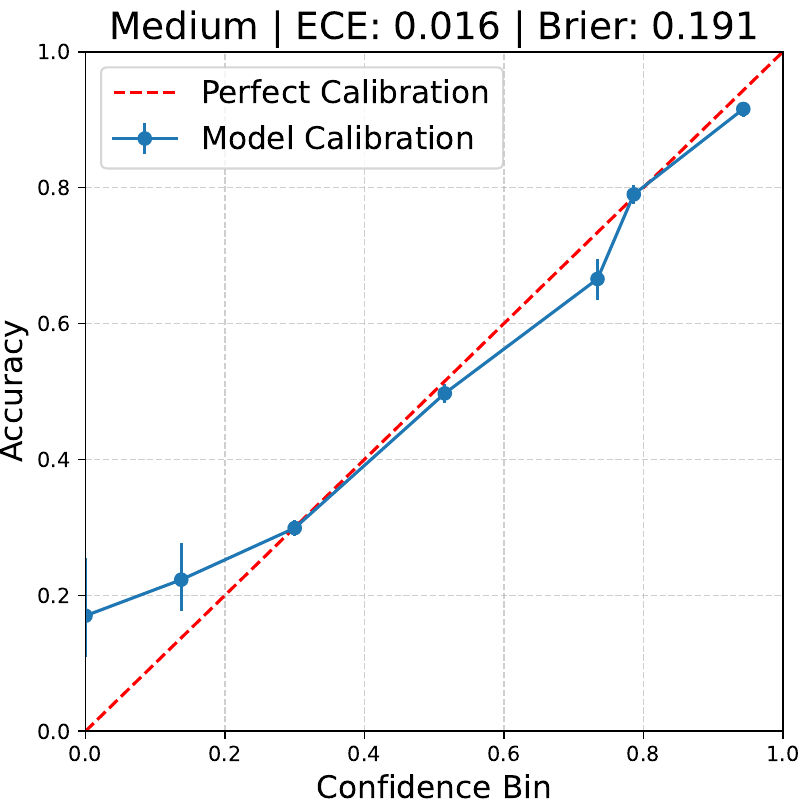}
        \captionsetup{justification=centering}
        \caption{\small \textbf{Medium}}
    \end{subfigure}
    \begin{subfigure}[b]{0.23\textwidth}
        \centering
        \includegraphics[width=\textwidth]{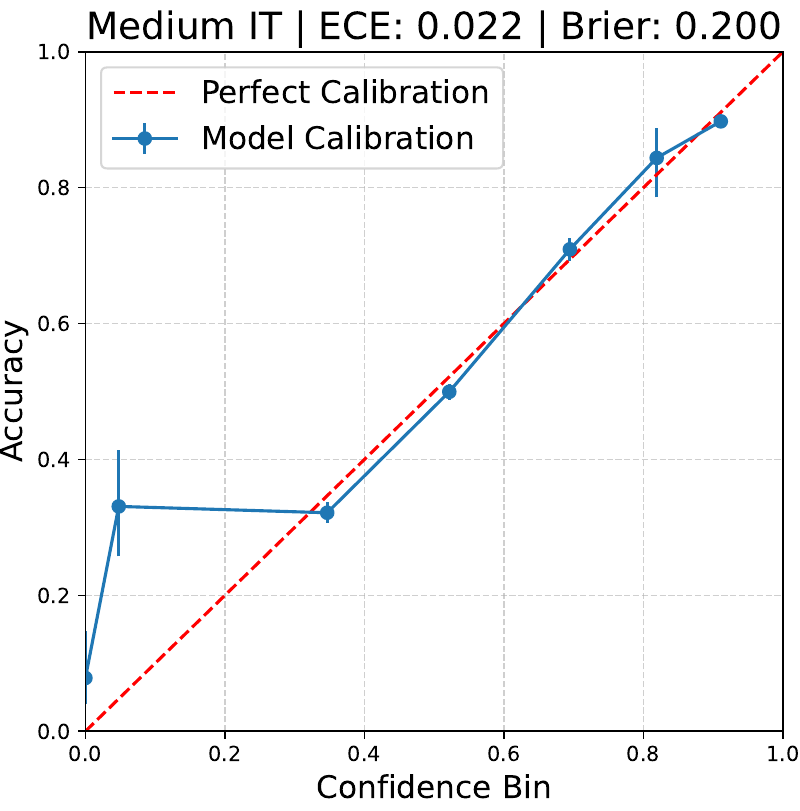}
        \captionsetup{justification=centering}
        \caption{\small \textbf{Medium-IT}}
    \end{subfigure}

    \caption{\textbf{AmbigQA Calibration Chart}: The \textbf{top-row} shows raw calibration scores at temperature=1.0 without any post-processing. The \textbf{bottom row} shows post-processed calibration scores with isotonic regression. In each plot, the x-axis is the $p_{model}(true)$ of the generated prediction (shown here as \texttt{Confidence Bin}) and y-axis is probability of that prediction being actually correct (shown here as \texttt{Accuracy}). Expected Calibration Error (ECE) and Brier Score are reported at the top of each plot.}
    \label{fig:ambig_qa_calibration_charts}
\end{figure}

\begin{figure}[htp]
    \centering
    \begin{subfigure}[b]{0.23\textwidth}
        \centering
        \includegraphics[width=\textwidth]{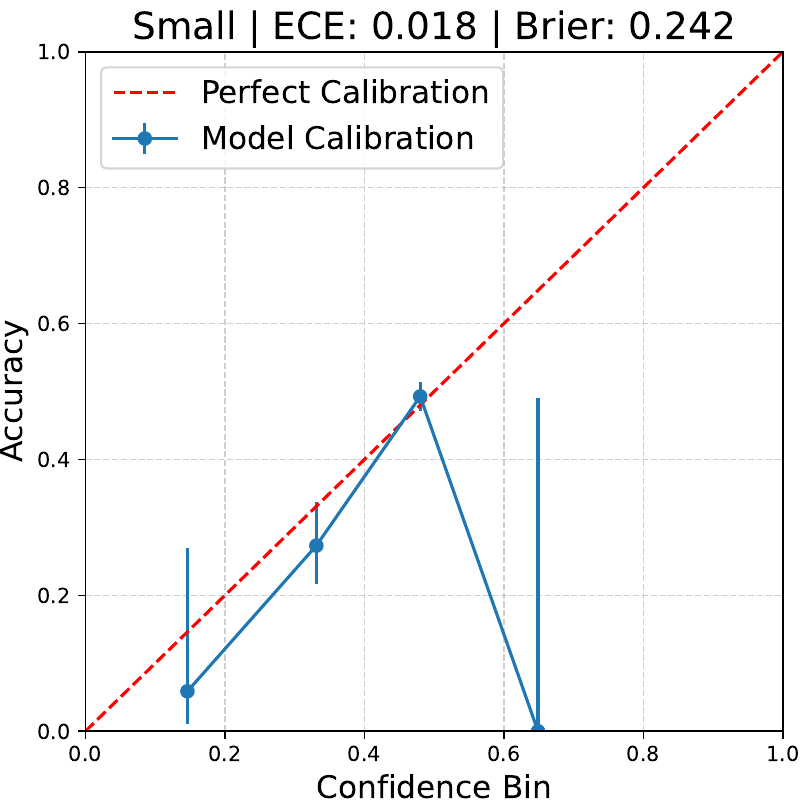}
    \end{subfigure}
    \begin{subfigure}[b]{0.23\textwidth}
        \centering
        \includegraphics[width=\textwidth]{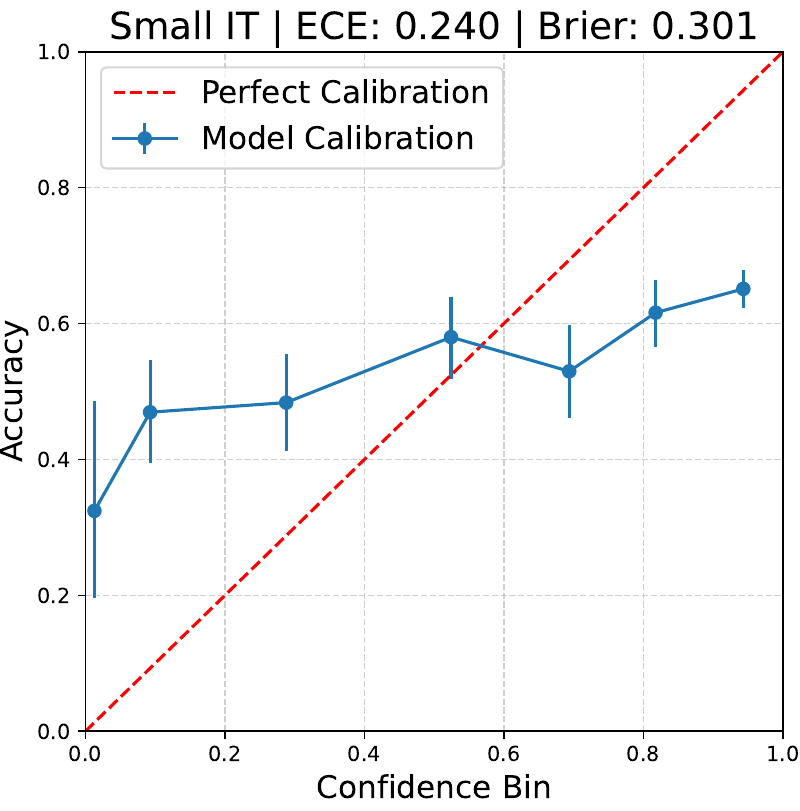}
    \end{subfigure}
    \begin{subfigure}[b]{0.23\textwidth}
        \centering
        \includegraphics[width=\textwidth]{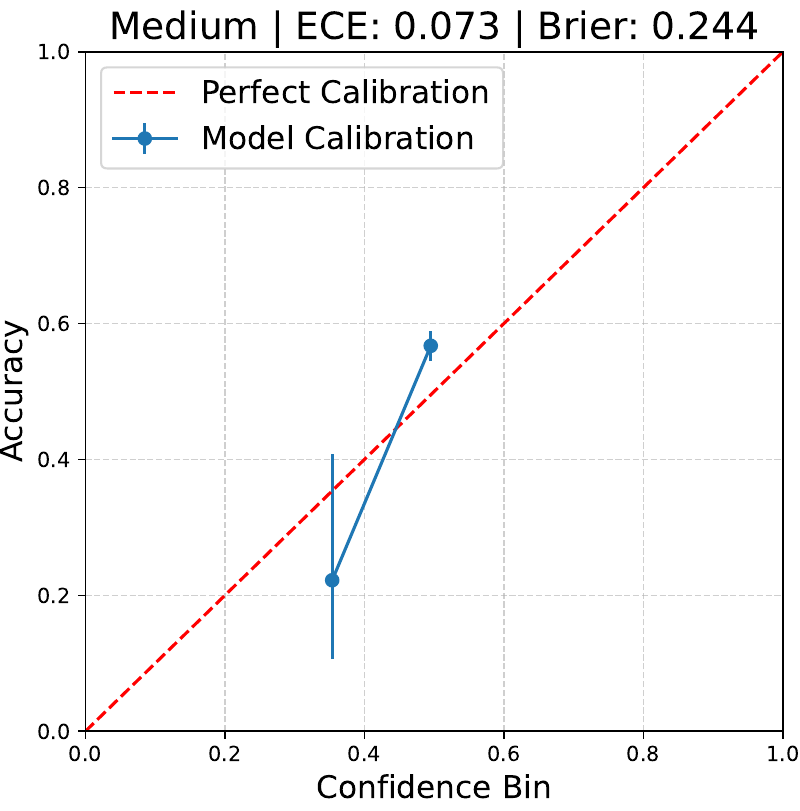}
    \end{subfigure}
    \begin{subfigure}[b]{0.23\textwidth}
        \centering
        \includegraphics[width=\textwidth]{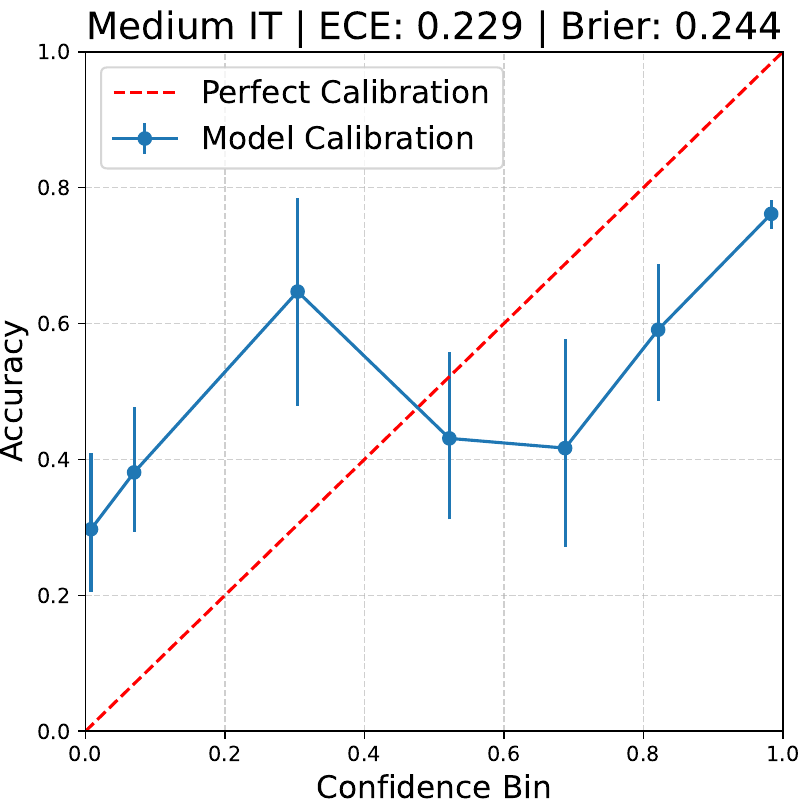}
    \end{subfigure}

    \begin{subfigure}[b]{0.23\textwidth}
        \centering
        \includegraphics[width=\textwidth]{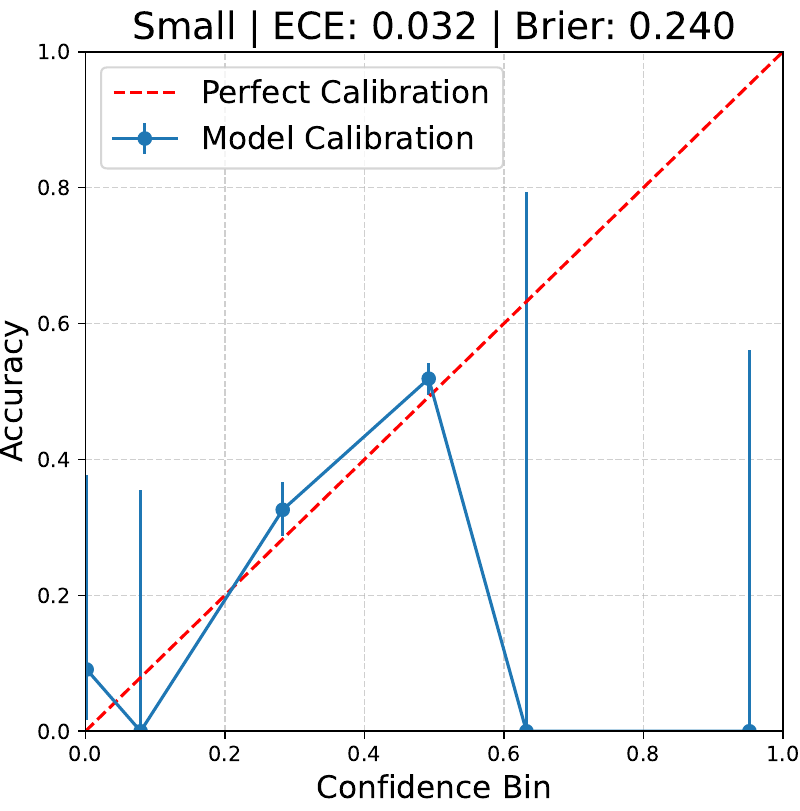}
        \captionsetup{justification=centering}
        \caption{\small \textbf{Small}}
    \end{subfigure}
    \begin{subfigure}[b]{0.23\textwidth}
        \centering
        \includegraphics[width=\textwidth]{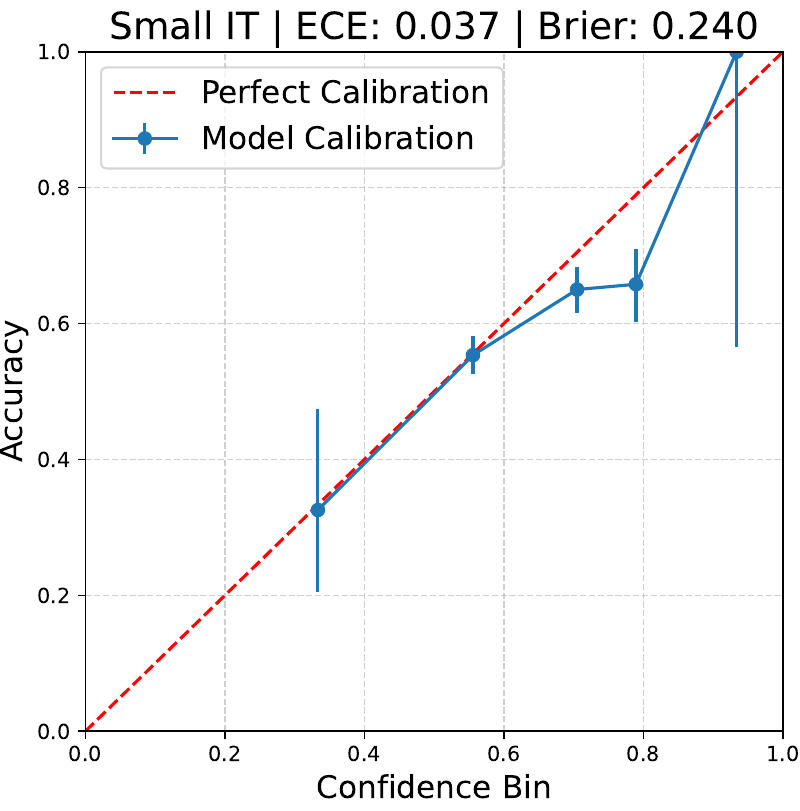}
        \captionsetup{justification=centering}
        \caption{\small \textbf{Small-IT}}
    \end{subfigure}
    \begin{subfigure}[b]{0.23\textwidth}
        \centering
        \includegraphics[width=\textwidth]{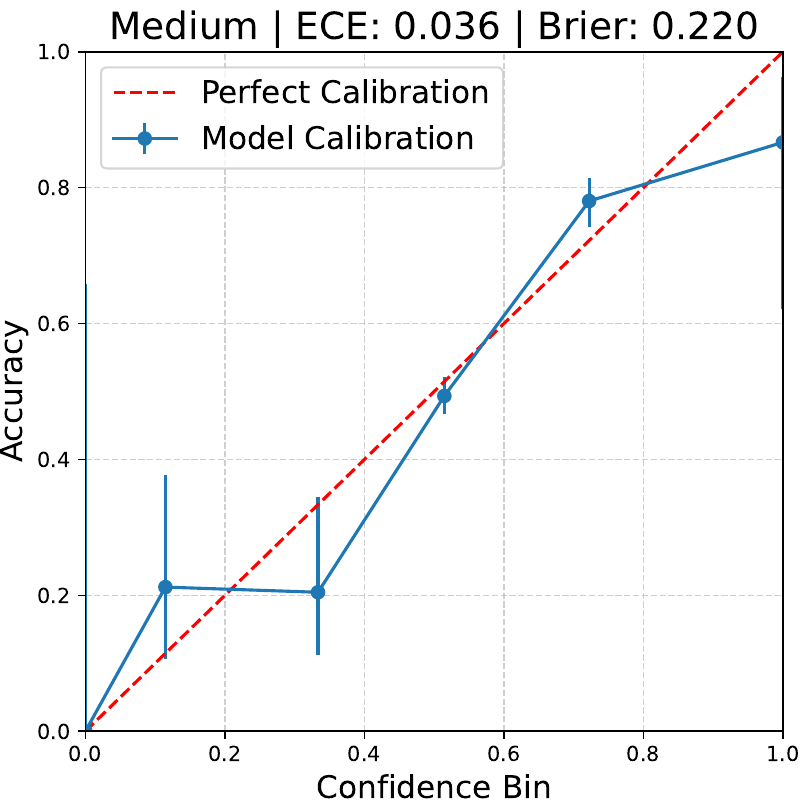}
        \captionsetup{justification=centering}
        \caption{\small \textbf{Medium}}
    \end{subfigure}
    \begin{subfigure}[b]{0.23\textwidth}
        \centering
        \includegraphics[width=\textwidth]{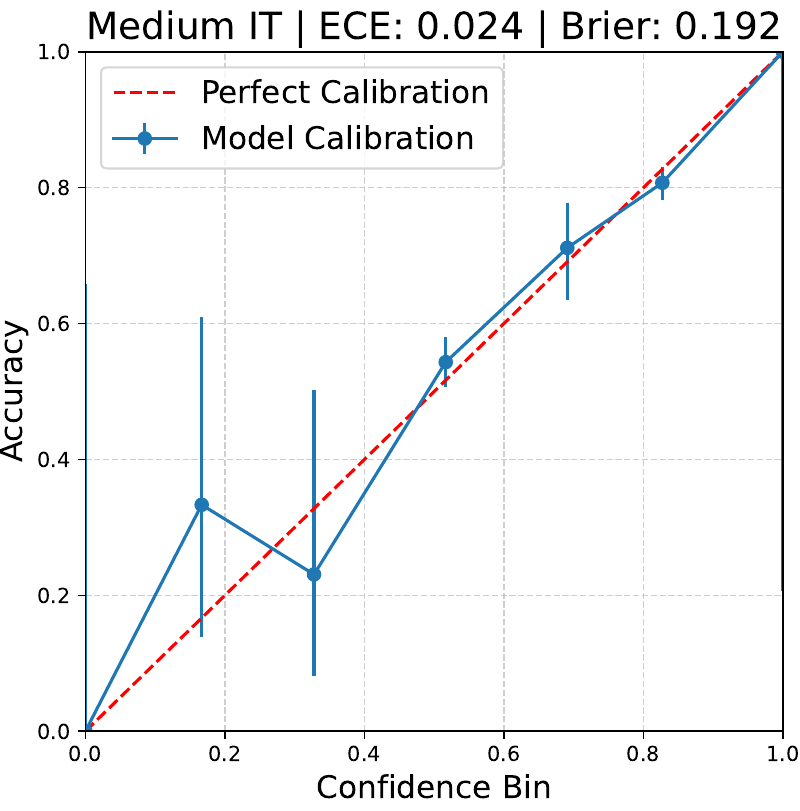}
        \captionsetup{justification=centering}
        \caption{\small \textbf{Medium-IT}}
    \end{subfigure}

    \caption{\textbf{TruthfulQA Calibration Chart}: The \textbf{top-row} shows raw calibration scores at temperature=1.0 without any post-processing. The \textbf{bottom row} shows post-processed calibration scores with isotonic regression. In each plot, the x-axis is the $p_{model}(true)$ of the generated prediction (shown here as \texttt{Confidence Bin}) and y-axis is probability of that prediction being actually correct (shown here as \texttt{Accuracy}). Expected Calibration Error (ECE) and Brier Score are reported at the top of each plot.}
    \label{fig:truthful_qa_calibration_charts}
\end{figure}


\begin{figure}[!t]
    \centering
    \begin{subfigure}[b]{0.32\textwidth}
        \centering
        \includegraphics[width=\textwidth]{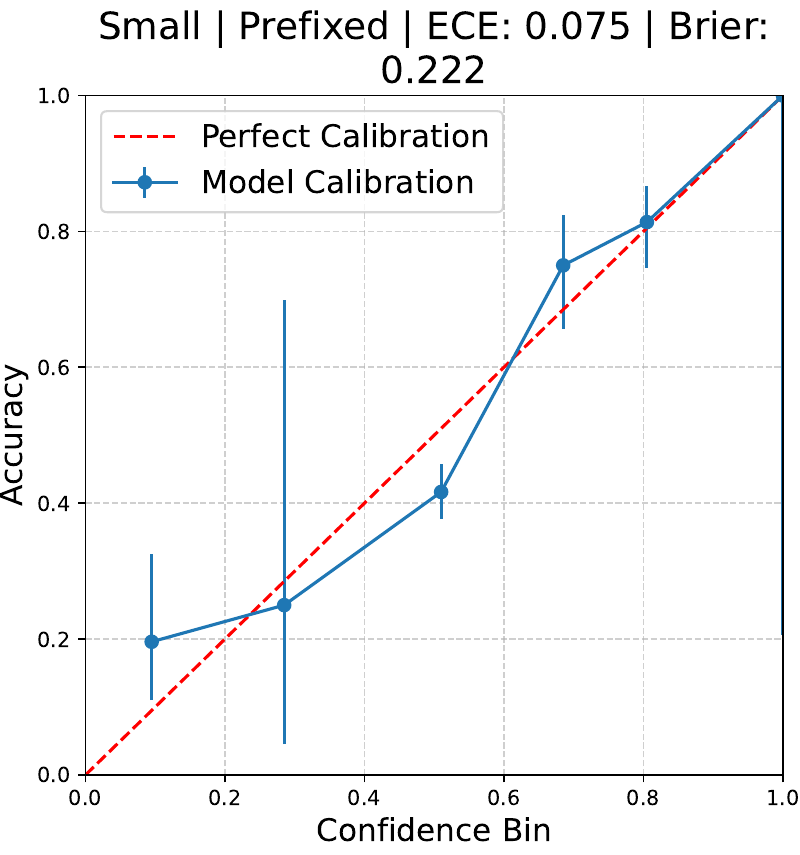}
    \end{subfigure}
    \begin{subfigure}[b]{0.32\textwidth}
        \centering
        \includegraphics[width=\textwidth]{Figures/finetuned/language_pred_prob_calibration_test_set_gemini_small_ambig_qa_postfixed.pdf}
    \end{subfigure}
    \begin{subfigure}[b]{0.32\textwidth}
        \centering
        \includegraphics[width=\textwidth]{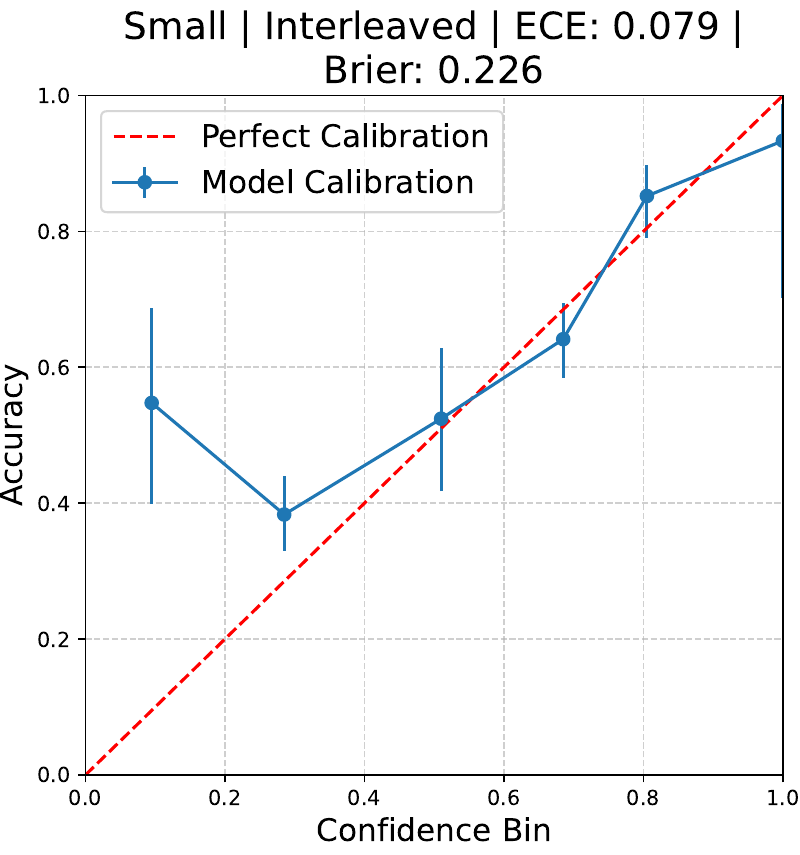}
    \end{subfigure}

    \begin{subfigure}[b]{0.32\textwidth}
        \centering
        \includegraphics[width=\textwidth]{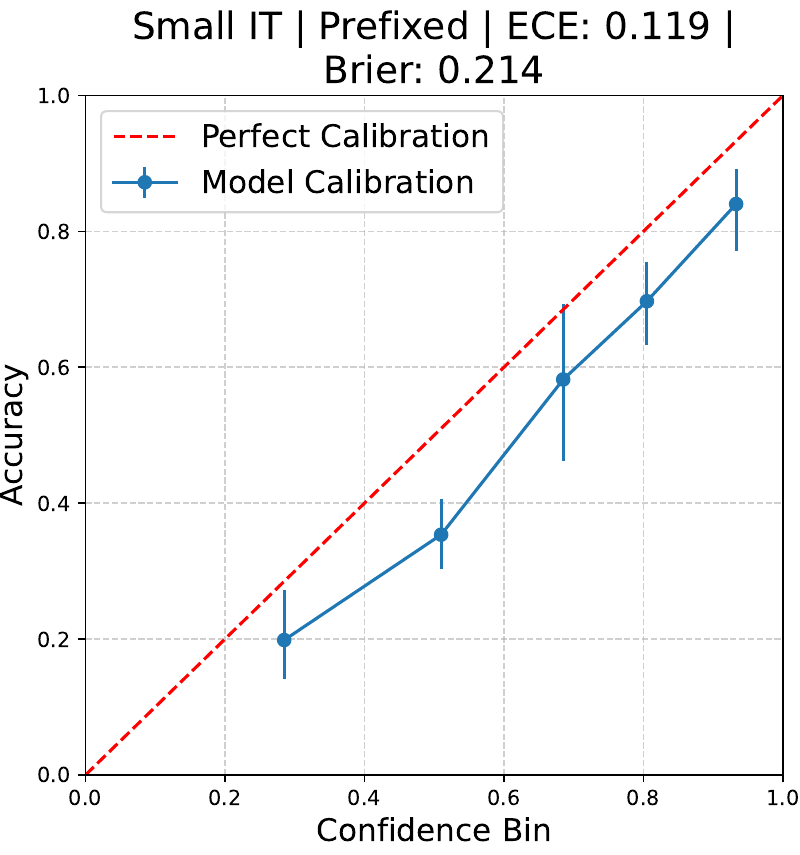}
    \end{subfigure}
    \begin{subfigure}[b]{0.32\textwidth}
        \centering
        \includegraphics[width=\textwidth]{Figures/finetuned/language_pred_prob_calibration_test_set_gemini_small_it_ambig_qa_postfixed.pdf}
    \end{subfigure}
    \begin{subfigure}[b]{0.32\textwidth}
        \centering
        \includegraphics[width=\textwidth]{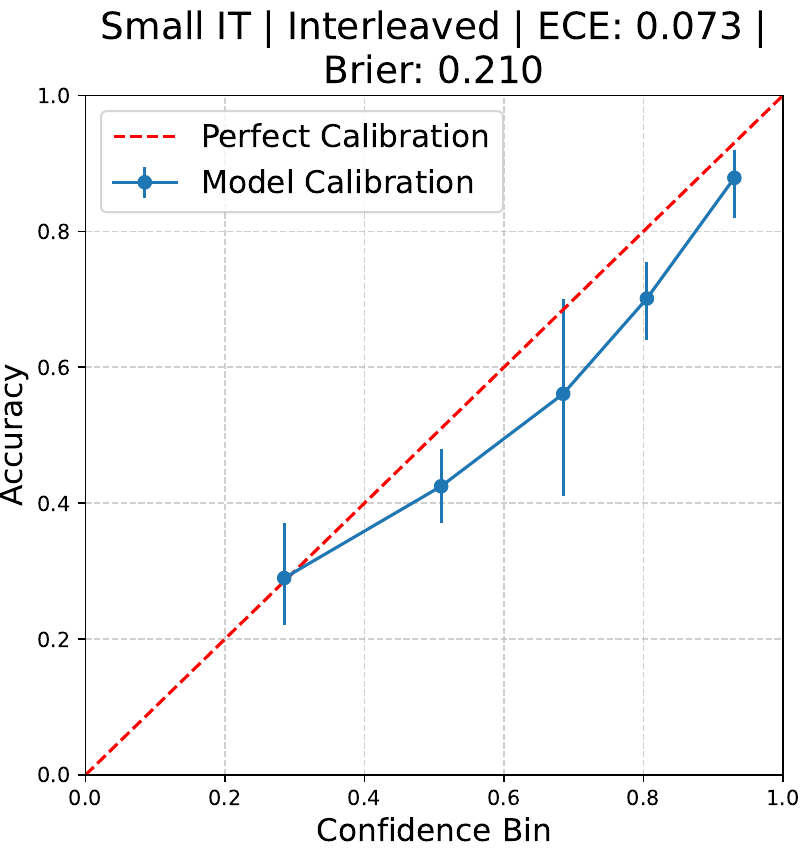}
    \end{subfigure}
    
    \begin{subfigure}[b]{0.32\textwidth}
        \centering
        \includegraphics[width=\textwidth]{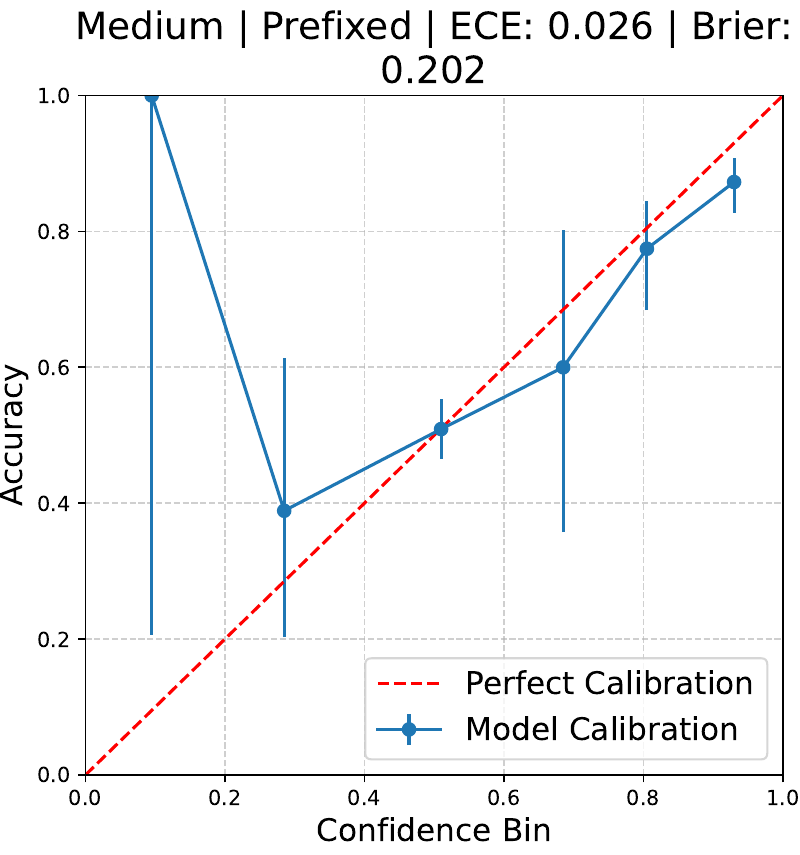}
        \captionsetup{justification=centering}
        \caption{\small \textbf{Prefixed}}
    \end{subfigure}
    \begin{subfigure}[b]{0.32\textwidth}
        \centering
        \includegraphics[width=\textwidth]{Figures/finetuned/language_pred_prob_calibration_test_set_gemini_medium_ambig_qa_postfixed.pdf}
        \captionsetup{justification=centering}
        \caption{\small \textbf{Postfixed}}
    \end{subfigure}
    \begin{subfigure}[b]{0.32\textwidth}
        \centering
        \includegraphics[width=\textwidth]{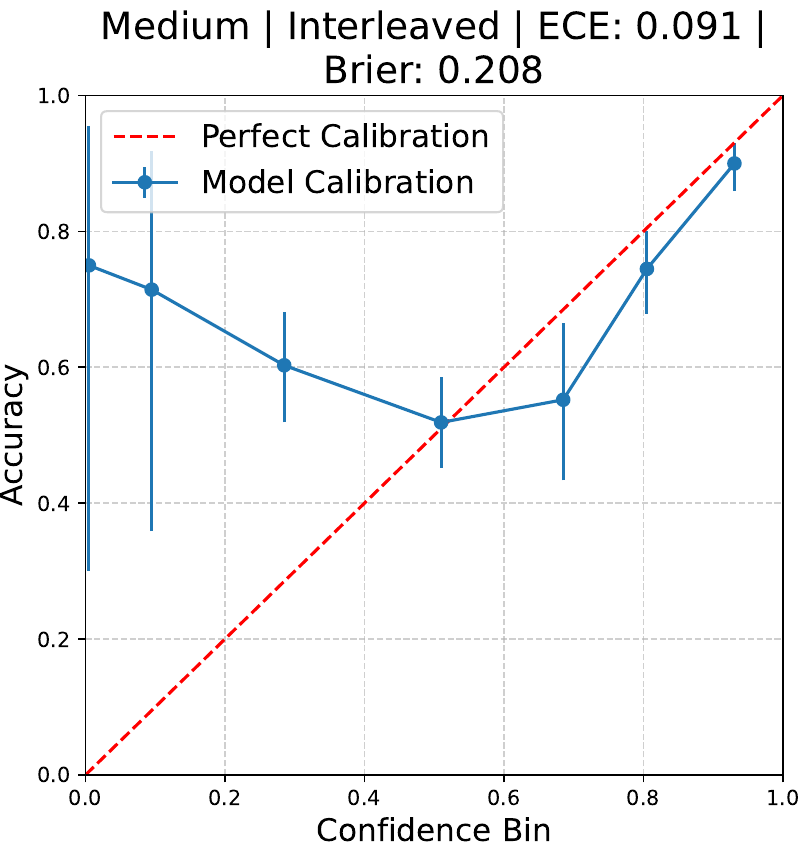}
        \captionsetup{justification=centering}
        \caption{\small \textbf{Interleaved}}
    \end{subfigure}

    \caption{\textbf{AmbigQA uncertainty augmentation method}: Top-row is \textbf{Gemini Small},  middle-row is \textbf{Gemini Small-IT} and the bottom row is \textbf{Gemini Medium}. The models generate \textbf{post-fixed} uncertainty expressions. The x-axis is the $p_{model}(true)$ obtained by converting the linguistic expression of uncertainty to a float using \Tabref{tab:expression_map} (shown here as \texttt{Confidence Bin}) and y-axis is probability of that prediction being actually correct (shown here as \texttt{Accuracy}). No post-processing is done on the $p_{model}(true)$. Expected Calibration Error (ECE) and Brier Score are reported at the top of each plot. The error bars show the variance of accuracy in each bin.}
    \label{fig:ambig_qa_augmentation_methods}
\end{figure}

\stop

\begin{figure*}[htp]
\captionsetup[subfigure]{justification=centering}
\centering
\begin{subfigure}[b]{0.45\textwidth}
    \includegraphics[width=\textwidth]{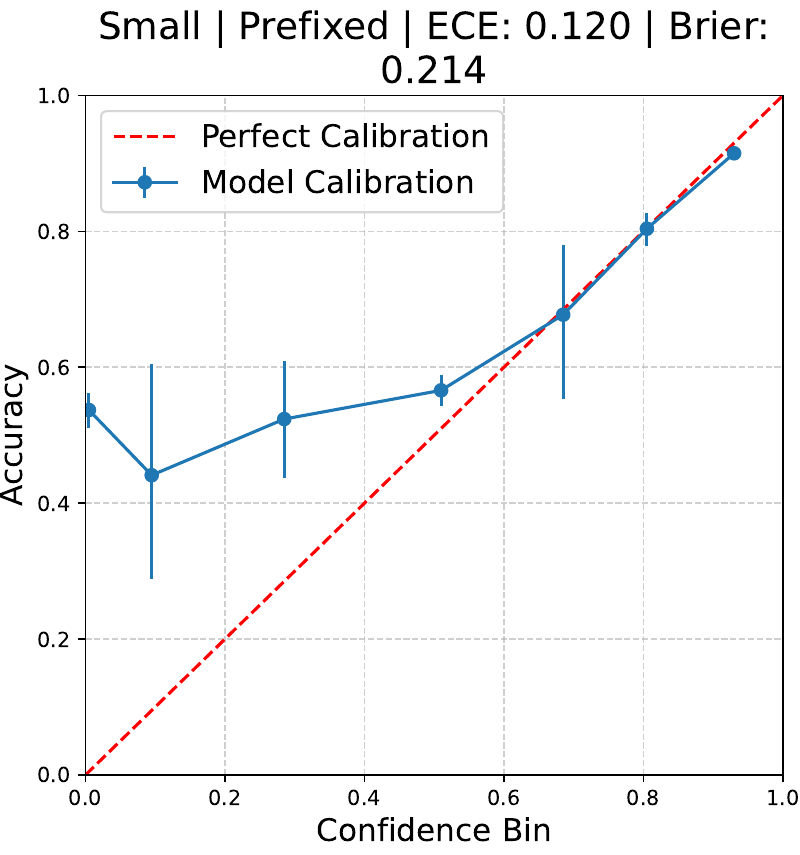}
    \caption{Prefixed}
    \label{fig:language_pred_prob_calibration_test_set_gemini_small_trivia_qa_prefixed}
\end{subfigure}\hfill
\begin{subfigure}[b]{0.45\textwidth}
    \includegraphics[width=\textwidth]{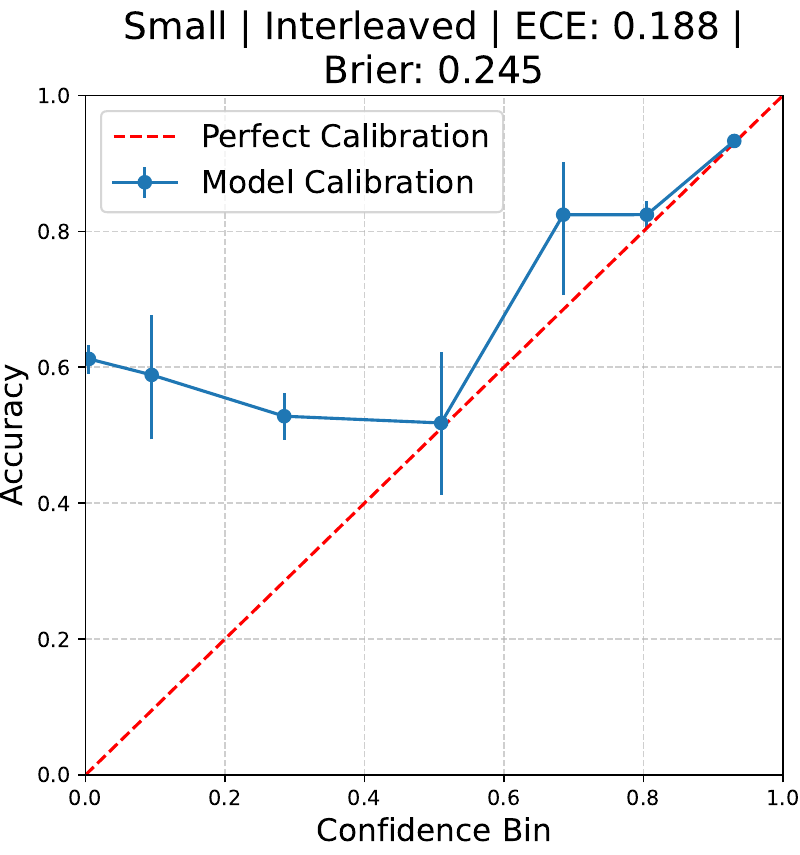}
    \caption{Prefixed Interleaved}
    \label{fig:language_pred_prob_calibration_test_set_gemini_small_trivia_qa_prefixed_interleaved}
\end{subfigure}

\begin{subfigure}[b]{0.45\textwidth}
    \includegraphics[width=\textwidth]{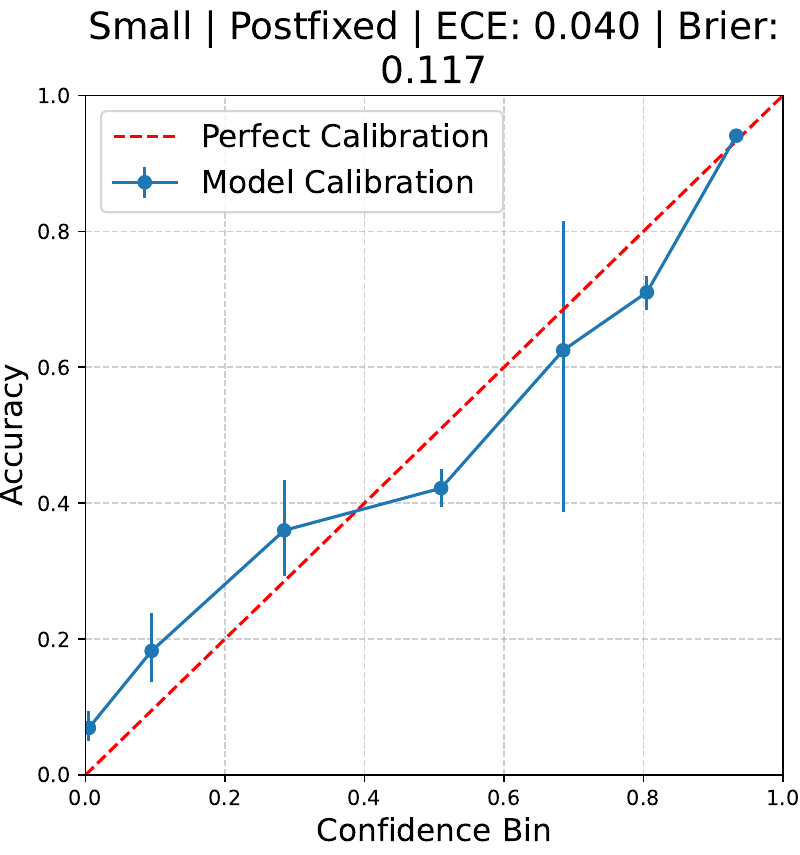}
    \caption{Postfixed}
    \label{fig:language_pred_prob_calibration_test_set_gemini_small_trivia_qa_postfixed}
\end{subfigure}\hfill

\caption{\textbf{Trivia QA Test Calibration Plots of Finetuned Model (Gemini Small, In Distribution)}: The x-axis is the $p_{model}(true)$ obtained by converting the linguistic uncertainty expression, in the prediction of the finetuned model, to a float (shown here as \texttt{Confidence Bin}) and y-axis is probability of that prediction being actually correct (shown here as \texttt{Accuracy}). No post-processing is done on the $p_{model}(true)$. Expected Calibration Error (ECE) and Brier Score are reported at the top of each plot.}
\label{fig:language_pred_prob_calibration_test_set_gemini_small_trivia_qa}
\end{figure*}

\begin{figure*}[htp]
\captionsetup[subfigure]{justification=centering}
\centering
\begin{subfigure}[b]{0.45\textwidth}
    \includegraphics[width=\textwidth]{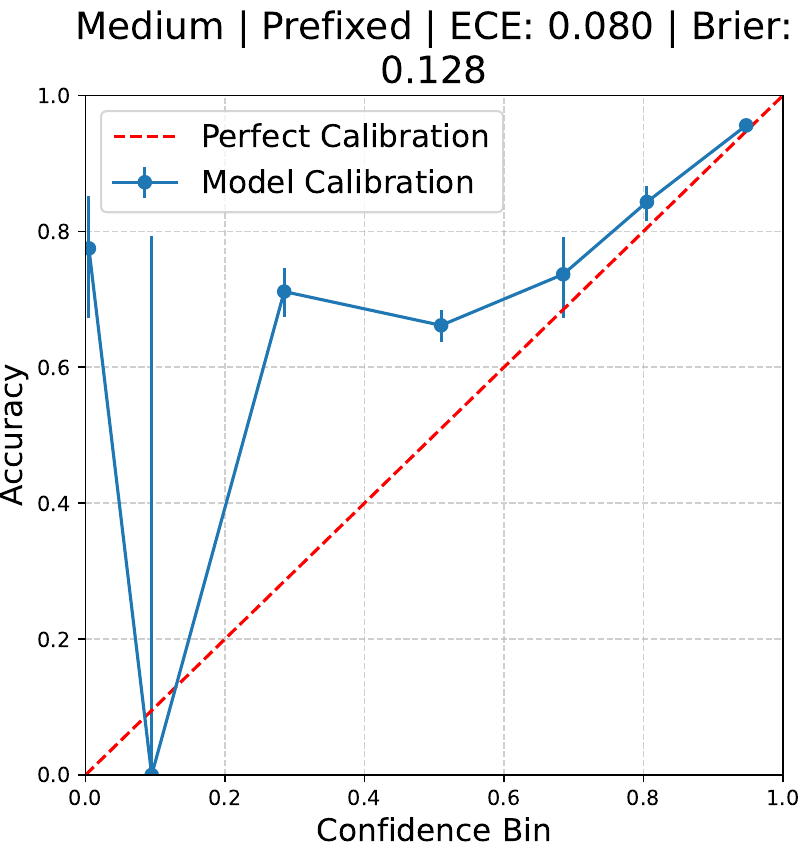}
    \caption{Prefixed}
    \label{fig:language_pred_prob_calibration_test_set_gemini_medium_trivia_qa_prefixed}
\end{subfigure}\hfill
\begin{subfigure}[b]{0.45\textwidth}
    \includegraphics[width=\textwidth]{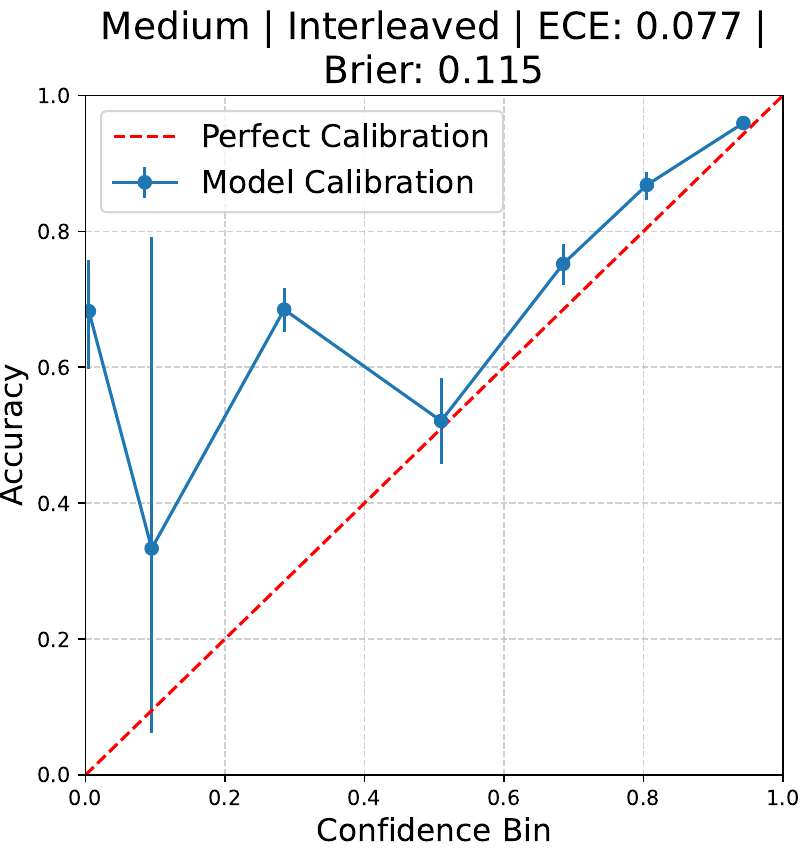}
    \caption{Prefixed Interleaved}
    \label{fig:language_pred_prob_calibration_test_set_gemini_medium_trivia_qa_prefixed_interleaved}
\end{subfigure}

\begin{subfigure}[b]{0.45\textwidth}
    \includegraphics[width=\textwidth]{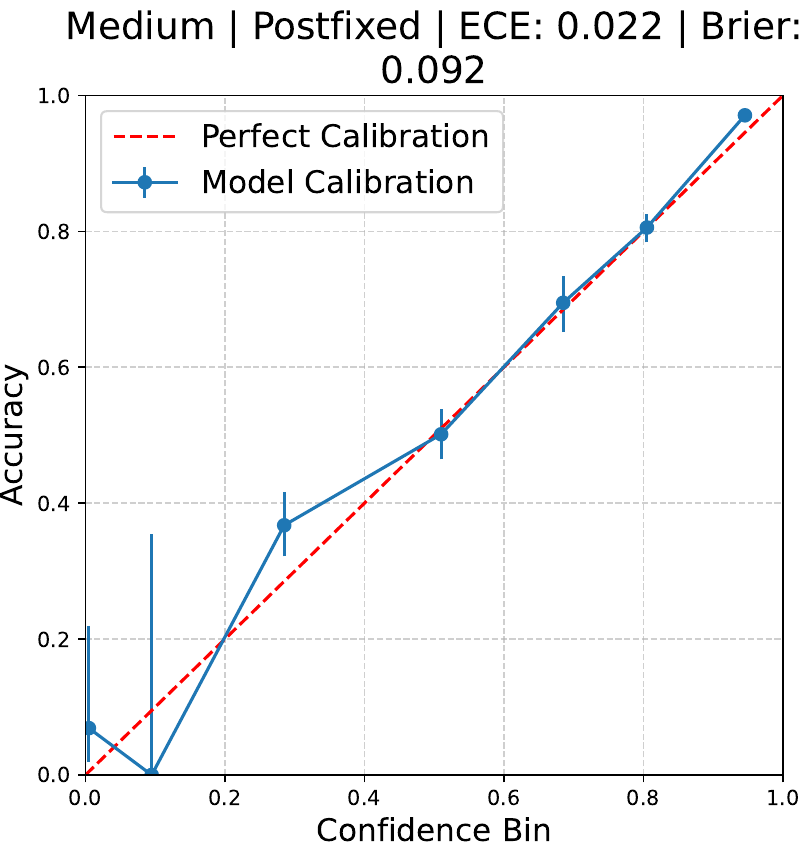}
    \caption{Postfixed}
    \label{fig:language_pred_prob_calibration_test_set_gemini_medium_trivia_qa_postfixed}
\end{subfigure}\hfill

\caption{\textbf{Trivia QA Test Calibration Plots of Finetuned Model (Gemini Medium, In Distribution)}: The x-axis is the $p_{model}(true)$ obtained by converting the linguistic uncertainty expression, in the prediction of the finetuned model, to a float (shown here as \texttt{Confidence Bin}) and y-axis is probability of that prediction being actually correct (shown here as \texttt{Accuracy}). No post-processing is done on the $p_{model}(true)$. Expected Calibration Error (ECE) and Brier Score are reported at the top of each plot.}
\label{fig:language_pred_prob_calibration_test_set_gemini_medium_trivia_qa}
\end{figure*}

\begin{figure*}[htp]
\captionsetup[subfigure]{justification=centering}
\centering
\begin{subfigure}[b]{0.45\textwidth}
    \includegraphics[width=\textwidth]{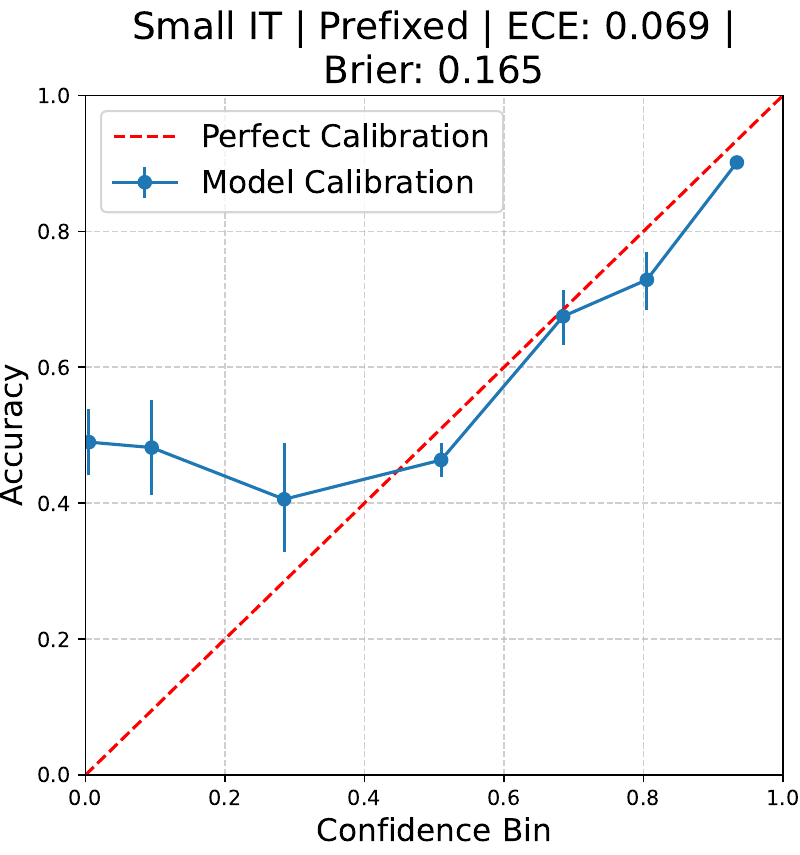}
    \caption{Prefixed}
    \label{fig:language_pred_prob_calibration_test_set_gemini_small_it_trivia_qa_prefixed}
\end{subfigure}\hfill
\begin{subfigure}[b]{0.45\textwidth}
    \includegraphics[width=\textwidth]{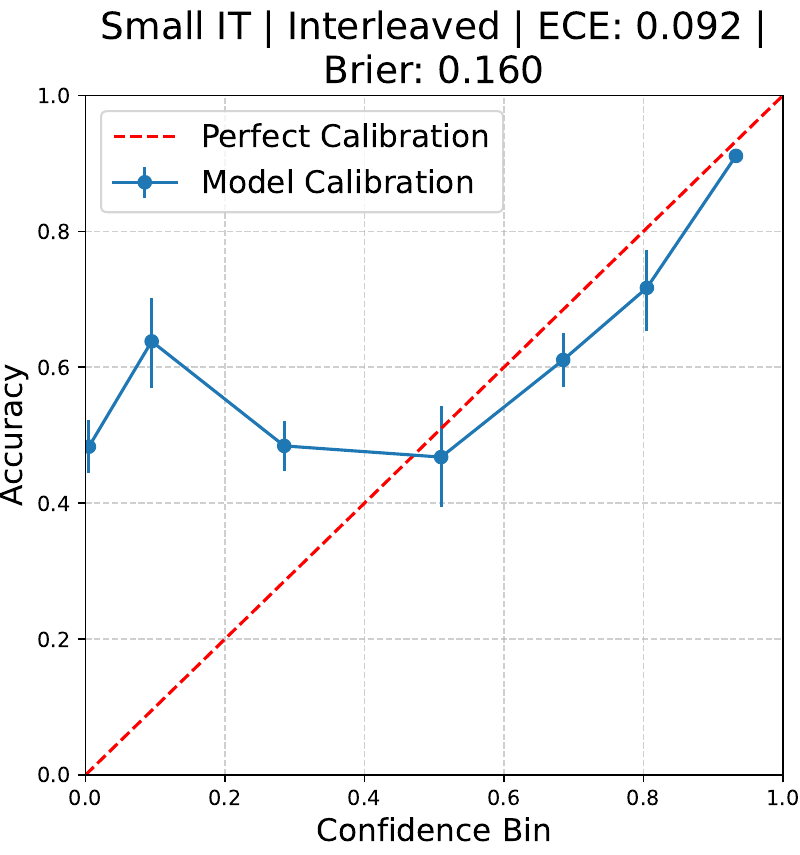}
    \caption{Prefixed Interleaved}
    \label{fig:language_pred_prob_calibration_test_set_gemini_small_it_trivia_qa_prefixed_interleaved}
\end{subfigure}

\begin{subfigure}[b]{0.45\textwidth}
    \includegraphics[width=\textwidth]{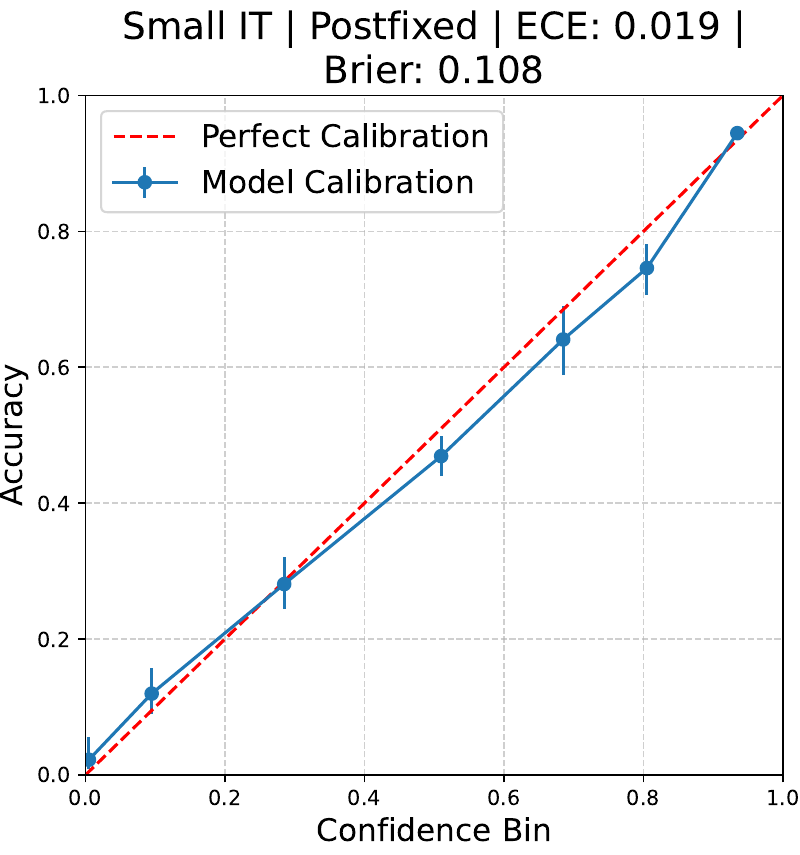}
    \caption{Postfixed}
    \label{fig:language_pred_prob_calibration_test_set_gemini_small_it_trivia_qa_postfixed}
\end{subfigure}\hfill

\caption{\textbf{Trivia QA Test Calibration Plots of Finetuned Model (Gemini IT Small, In Distribution)}: The x-axis is the $p_{model}(true)$ obtained by converting the linguistic uncertainty expression, in the prediction of the finetuned model, to a float (shown here as \texttt{Confidence Bin}) and y-axis is probability of that prediction being actually correct (shown here as \texttt{Accuracy}). No post-processing is done on the $p_{model}(true)$. Expected Calibration Error (ECE) and Brier Score are reported at the top of each plot.}
\label{fig:language_pred_prob_calibration_test_set_gemini_it_small_trivia_qa}
\end{figure*}

\begin{figure*}[htp]
\captionsetup[subfigure]{justification=centering}
\centering
\begin{subfigure}[b]{0.45\textwidth}
    \includegraphics[width=\textwidth]{Figures/finetuned/language_pred_prob_calibration_test_set_gemini_medium_it_trivia_qa_prefixed.pdf}
    \caption{Prefixed}
    \label{fig:language_pred_prob_calibration_test_set_gemini_medium_it_trivia_qa_prefixed}
\end{subfigure}\hfill
\begin{subfigure}[b]{0.45\textwidth}
    \includegraphics[width=\textwidth]{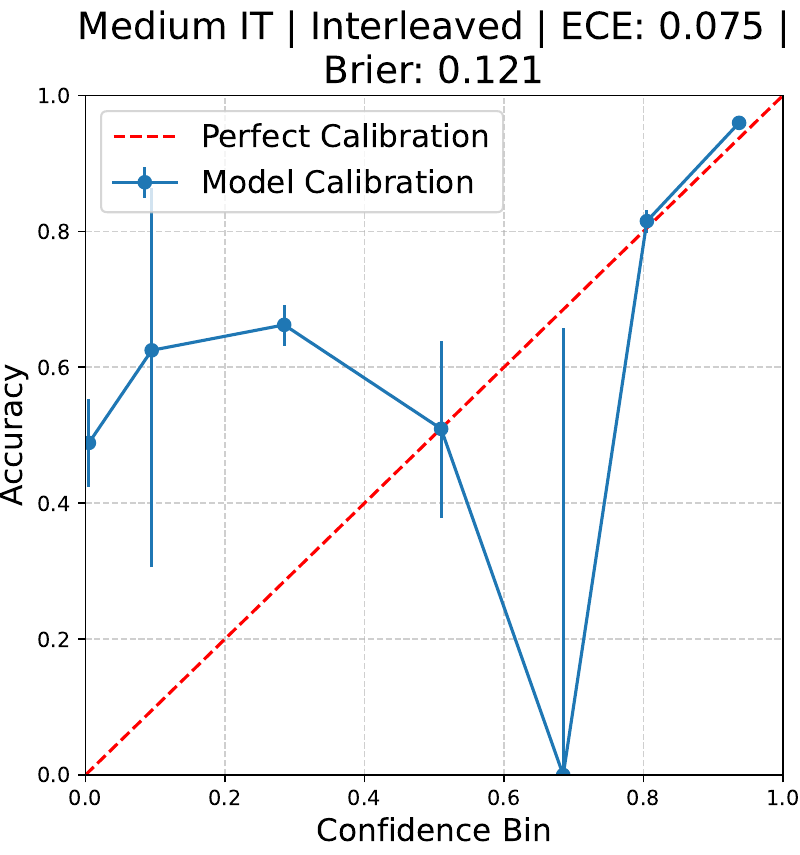}
    \caption{Prefixed Interleaved}
    \label{fig:language_pred_prob_calibration_test_set_gemini_medium_it_trivia_qa_prefixed_interleaved}
\end{subfigure}

\begin{subfigure}[b]{0.45\textwidth}
    \includegraphics[width=\textwidth]{Figures/finetuned/language_pred_prob_calibration_test_set_gemini_medium_it_trivia_qa_postfixed.pdf}
    \caption{Postfixed}
    \label{fig:language_pred_prob_calibration_test_set_gemini_medium_it_trivia_qa_postfixed}
\end{subfigure}\hfill

\caption{\textbf{Trivia QA Test Calibration Plots of Finetuned Model (Gemini IT Medium, In Distribution)}: The x-axis is the $p_{model}(true)$ obtained by converting the linguistic uncertainty expression, in the prediction of the finetuned model, to a float (shown here as \texttt{Confidence Bin}) and y-axis is probability of that prediction being actually correct (shown here as \texttt{Accuracy}). No post-processing is done on the $p_{model}(true)$. Expected Calibration Error (ECE) and Brier Score are reported at the top of each plot.}
\label{fig:language_pred_prob_calibration_test_set_gemini_it_medium_trivia_qa}
\end{figure*}


\begin{figure*}[htp]
\captionsetup[subfigure]{justification=centering}
\centering
\begin{subfigure}[b]{0.45\textwidth}
    \includegraphics[width=\textwidth]{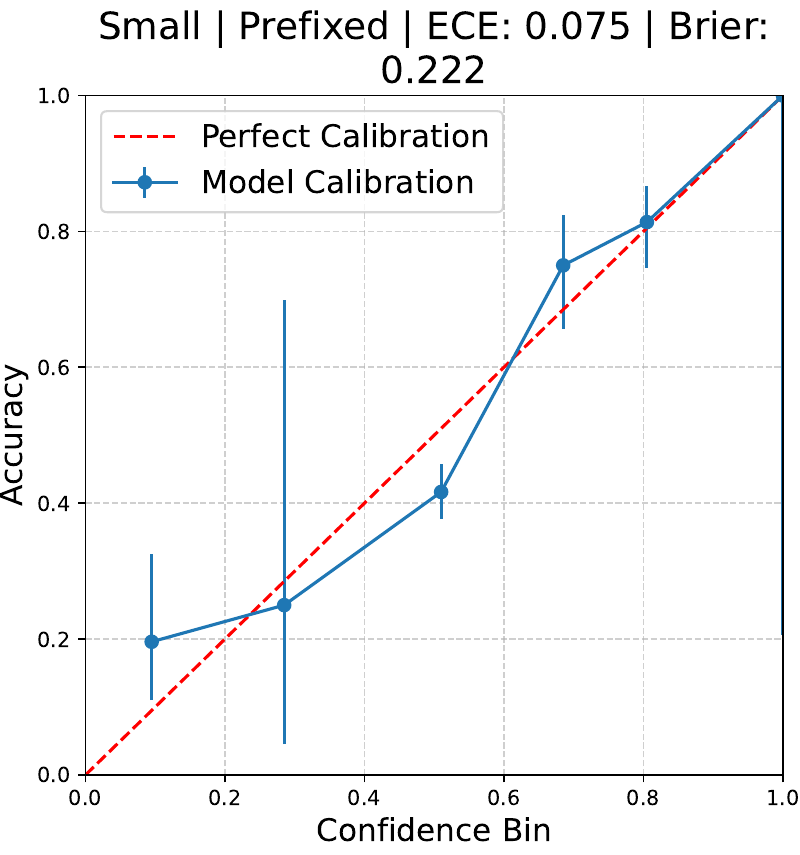}
    \caption{Prefixed}
    \label{fig:language_pred_prob_calibration_test_set_gemini_small_ambig_qa_prefixed}
\end{subfigure}\hfill
\begin{subfigure}[b]{0.45\textwidth}
    \includegraphics[width=\textwidth]{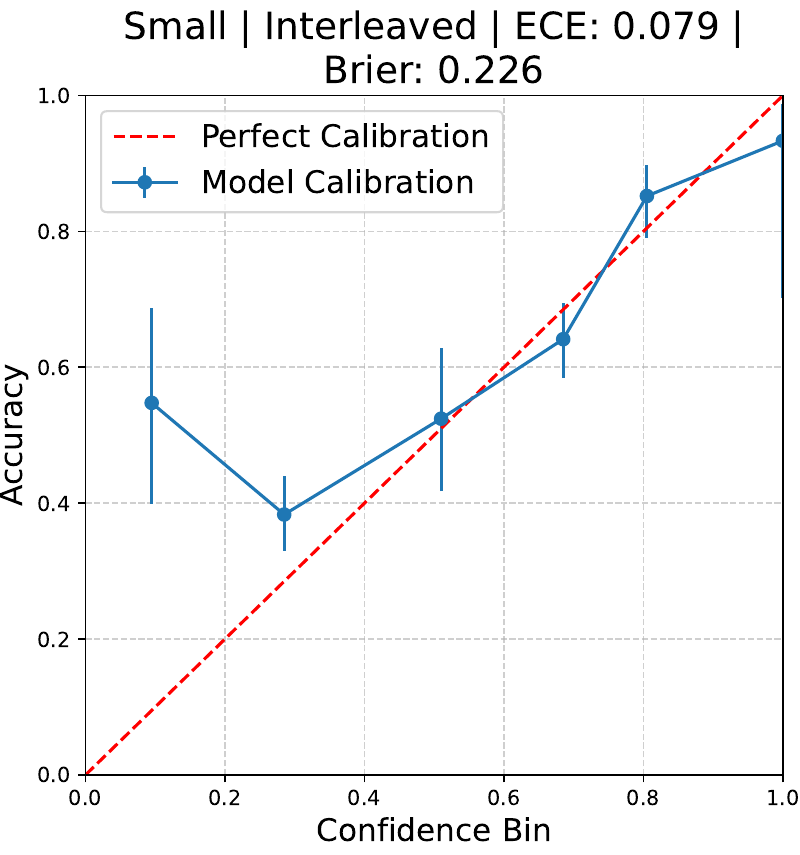}
    \caption{Prefixed Interleaved}
    \label{fig:language_pred_prob_calibration_test_set_gemini_small_ambig_qa_prefixed_interleaved}
\end{subfigure}

\begin{subfigure}[b]{0.45\textwidth}
    \includegraphics[width=\textwidth]{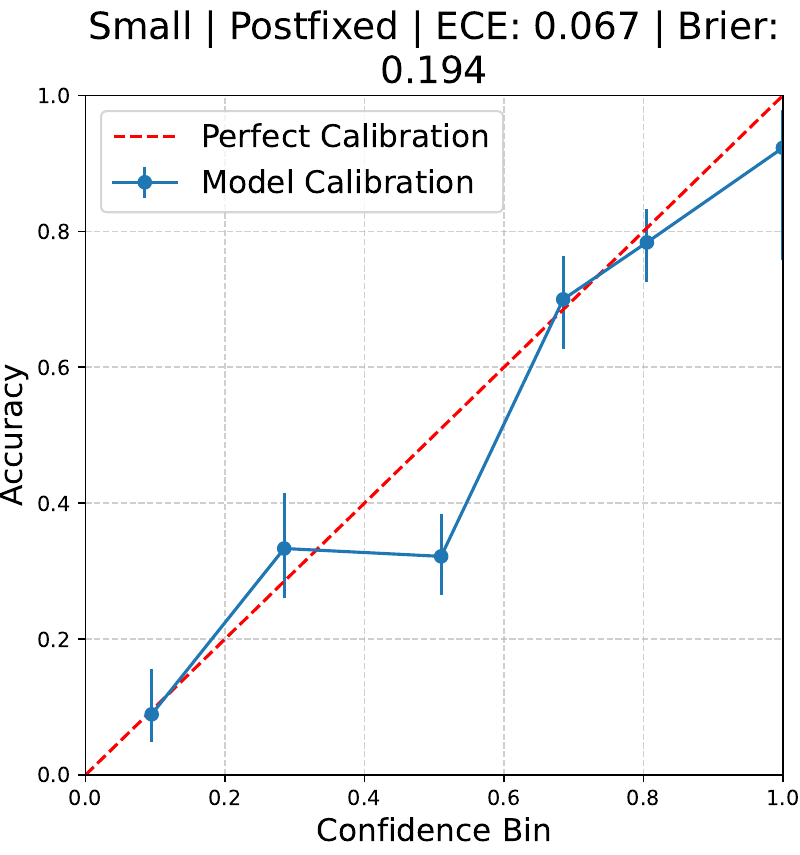}
    \caption{Postfixed}
    \label{fig:language_pred_prob_calibration_test_set_gemini_small_ambig_qa_postfixed}
\end{subfigure}\hfill

\caption{\textbf{Ambig QA Test Calibration Plots of Finetuned Model (Gemini Small, In Distribution)}: The x-axis is the $p_{model}(true)$ obtained by converting the linguistic uncertainty expression, in the prediction of the finetuned model, to a float (shown here as \texttt{Confidence Bin}) and y-axis is probability of that prediction being actually correct (shown here as \texttt{Accuracy}). No post-processing is done on the $p_{model}(true)$. Expected Calibration Error (ECE) and Brier Score are reported at the top of each plot.}
\label{fig:language_pred_prob_calibration_test_set_gemini_small_ambig_qa}
\end{figure*}

\begin{figure*}[htp]
\captionsetup[subfigure]{justification=centering}
\centering
\begin{subfigure}[b]{0.45\textwidth}
    \includegraphics[width=\textwidth]{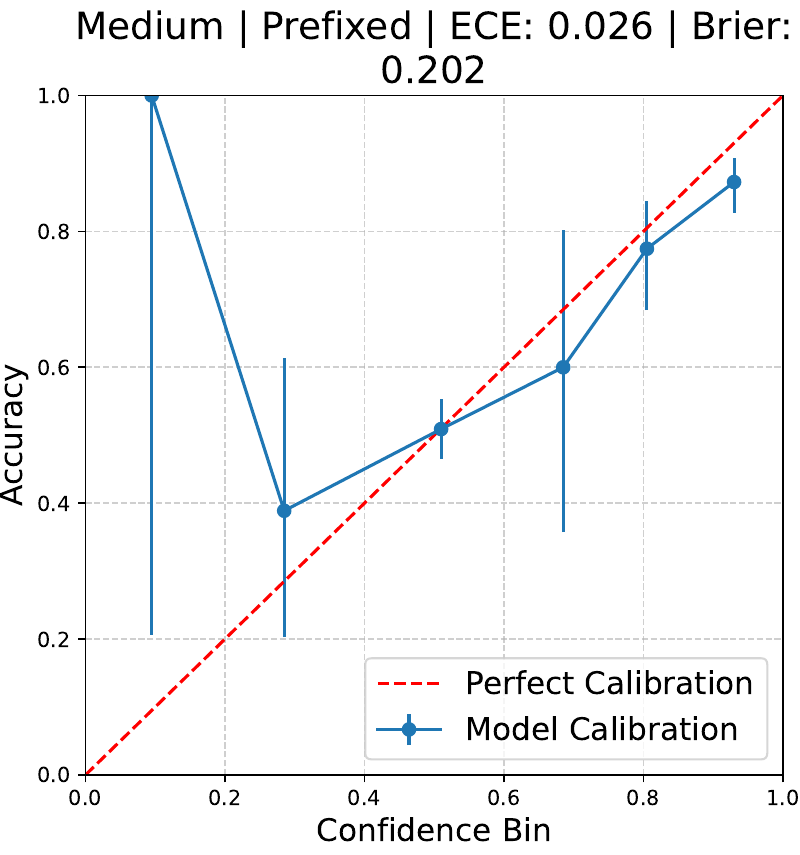}
    \caption{Prefixed}
    \label{fig:language_pred_prob_calibration_test_set_gemini_medium_ambig_qa_prefixed}
\end{subfigure}\hfill
\begin{subfigure}[b]{0.45\textwidth}
    \includegraphics[width=\textwidth]{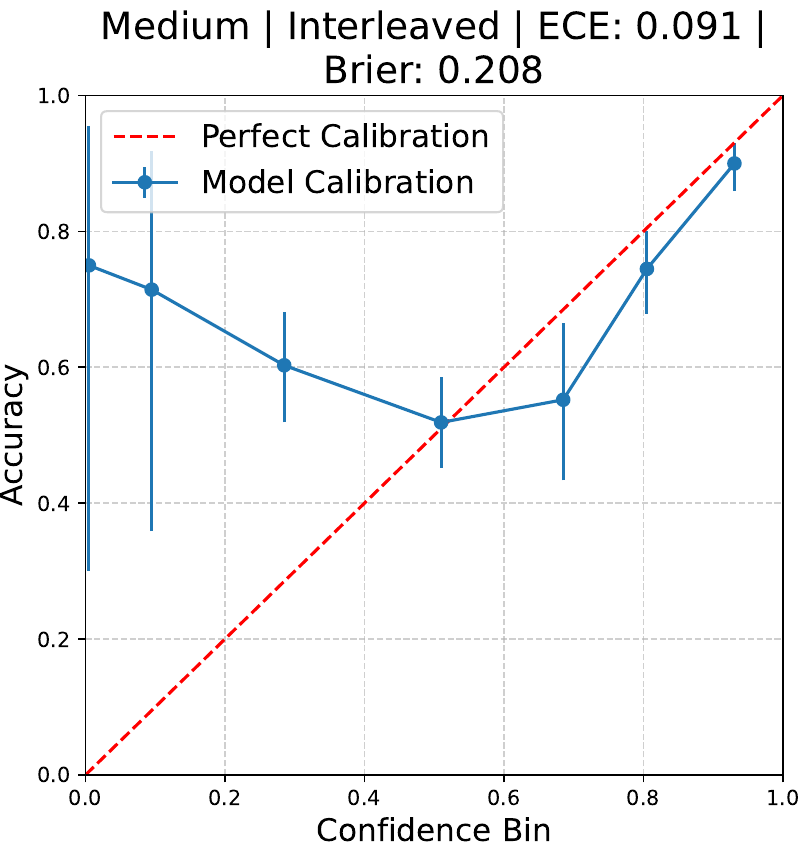}
    \caption{Prefixed Interleaved}
    \label{fig:language_pred_prob_calibration_test_set_gemini_medium_ambig_qa_prefixed_interleaved}
\end{subfigure}

\begin{subfigure}[b]{0.45\textwidth}
    \includegraphics[width=\textwidth]{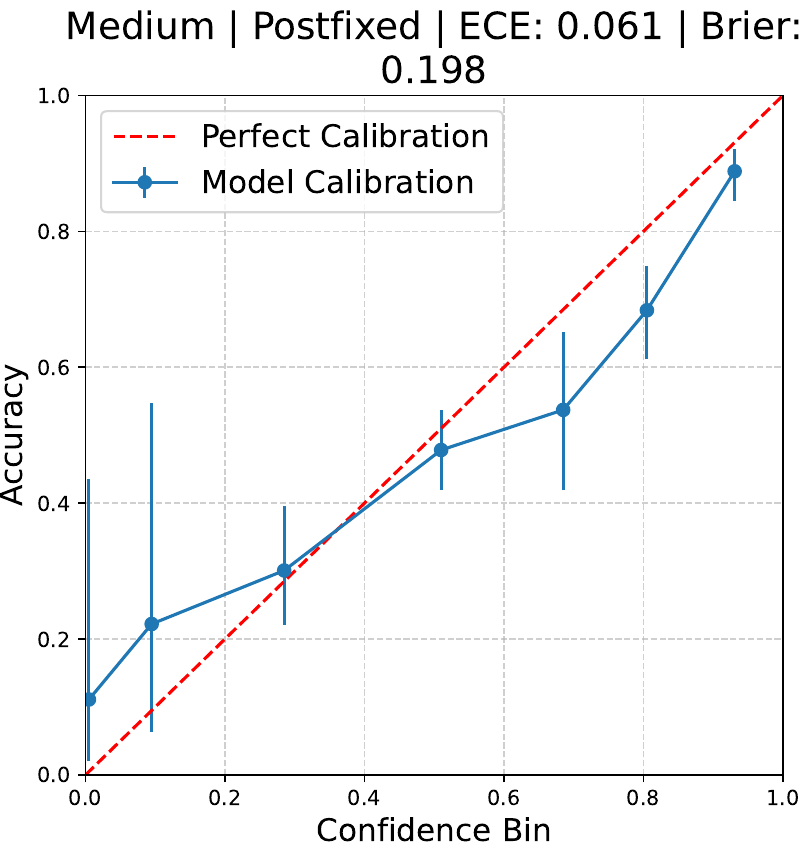}
    \caption{Postfixed}
    \label{fig:language_pred_prob_calibration_test_set_gemini_medium_ambig_qa_postfixed}
\end{subfigure}\hfill

\caption{\textbf{Ambig QA Test Calibration Plots of Finetuned Model (Gemini Medium, In Distribution)}: The x-axis is the $p_{model}(true)$ obtained by converting the linguistic uncertainty expression, in the prediction of the finetuned model, to a float (shown here as \texttt{Confidence Bin}) and y-axis is probability of that prediction being actually correct (shown here as \texttt{Accuracy}). No post-processing is done on the $p_{model}(true)$. Expected Calibration Error (ECE) and Brier Score are reported at the top of each plot.}
\label{fig:language_pred_prob_calibration_test_set_gemini_medium_ambig_qa}
\end{figure*}

\begin{figure*}[htp]
\captionsetup[subfigure]{justification=centering}
\centering
\begin{subfigure}[b]{0.45\textwidth}
    \includegraphics[width=\textwidth]{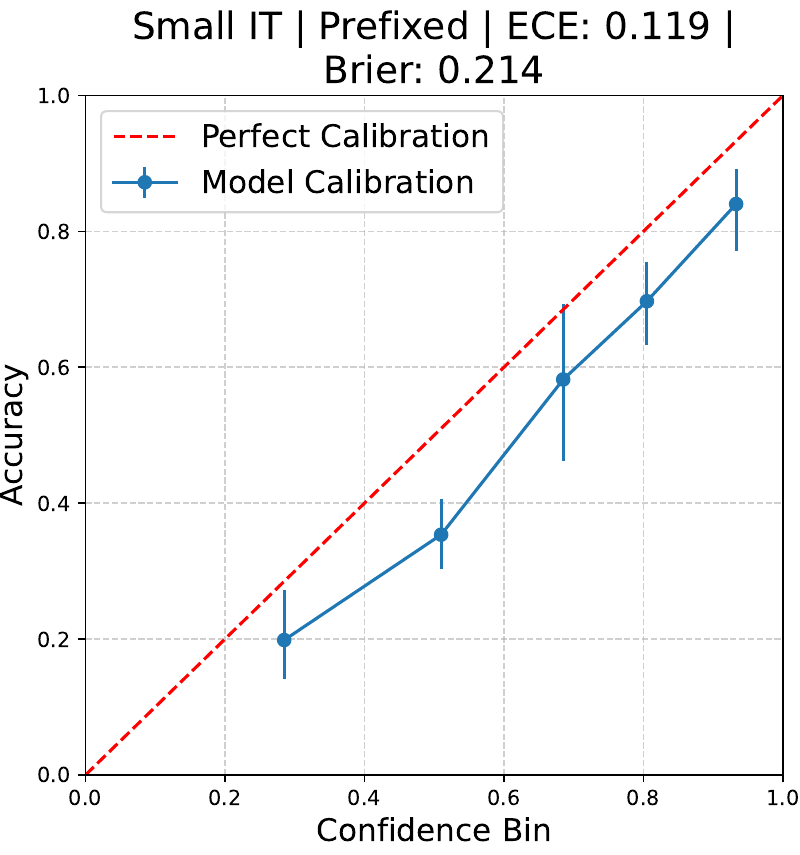}
    \caption{Prefixed}
    \label{fig:language_pred_prob_calibration_test_set_gemini_small_it_ambig_qa_prefixed}
\end{subfigure}\hfill
\begin{subfigure}[b]{0.45\textwidth}
    \includegraphics[width=\textwidth]{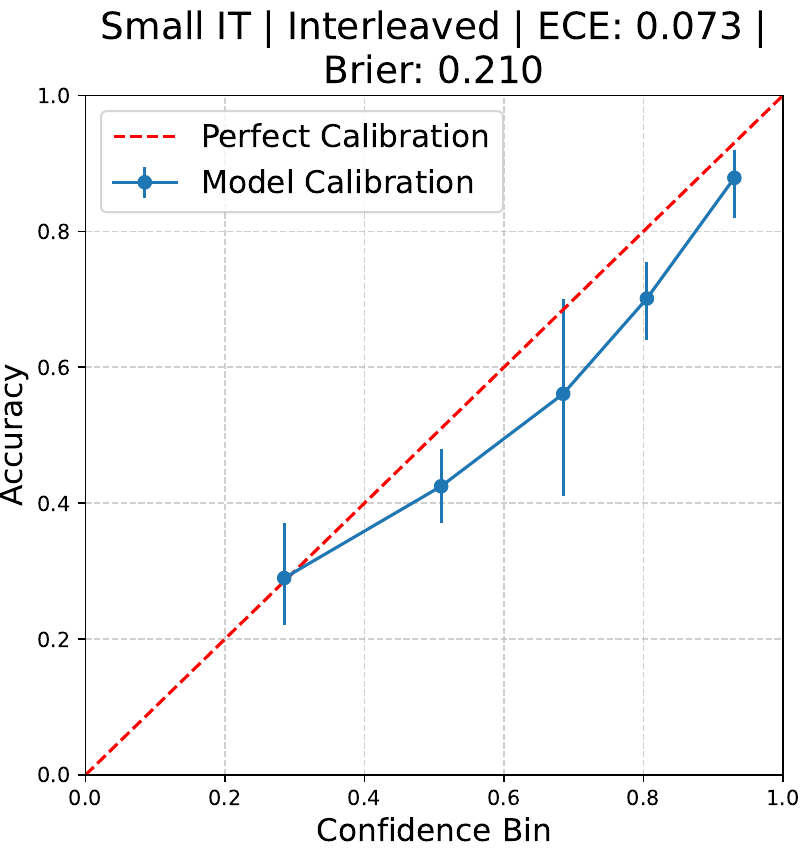}
    \caption{Prefixed Interleaved}
    \label{fig:language_pred_prob_calibration_test_set_gemini_small_it_ambig_qa_prefixed_interleaved}
\end{subfigure}

\begin{subfigure}[b]{0.45\textwidth}
    \includegraphics[width=\textwidth]{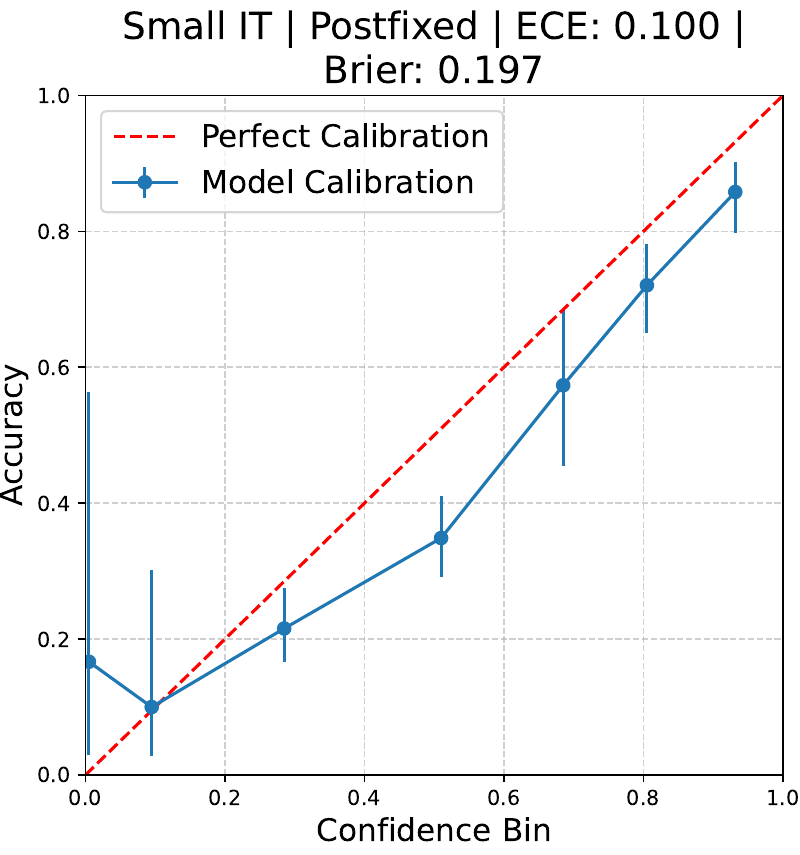}
    \caption{Postfixed}
    \label{fig:language_pred_prob_calibration_test_set_gemini_small_it_ambig_qa_postfixed}
\end{subfigure}\hfill

\caption{\textbf{Ambig QA Test Calibration Plots of Finetuned Model (Gemini IT Small, In Distribution)}: The x-axis is the $p_{model}(true)$ obtained by converting the linguistic uncertainty expression, in the prediction of the finetuned model, to a float (shown here as \texttt{Confidence Bin}) and y-axis is probability of that prediction being actually correct (shown here as \texttt{Accuracy}). No post-processing is done on the $p_{model}(true)$. Expected Calibration Error (ECE) and Brier Score are reported at the top of each plot.}
\label{fig:language_pred_prob_calibration_test_set_gemini_it_small_ambig_qa}
\end{figure*}

\begin{figure*}[htp]
\captionsetup[subfigure]{justification=centering}
\centering
\begin{subfigure}[b]{0.45\textwidth}
    \includegraphics[width=\textwidth]{Figures/finetuned/language_pred_prob_calibration_test_set_gemini_medium_it_ambig_qa_prefixed.pdf}
    \caption{Prefixed}
    \label{fig:language_pred_prob_calibration_test_set_gemini_medium_it_ambig_qa_prefixed}
\end{subfigure}\hfill
\begin{subfigure}[b]{0.45\textwidth}
    \includegraphics[width=\textwidth]{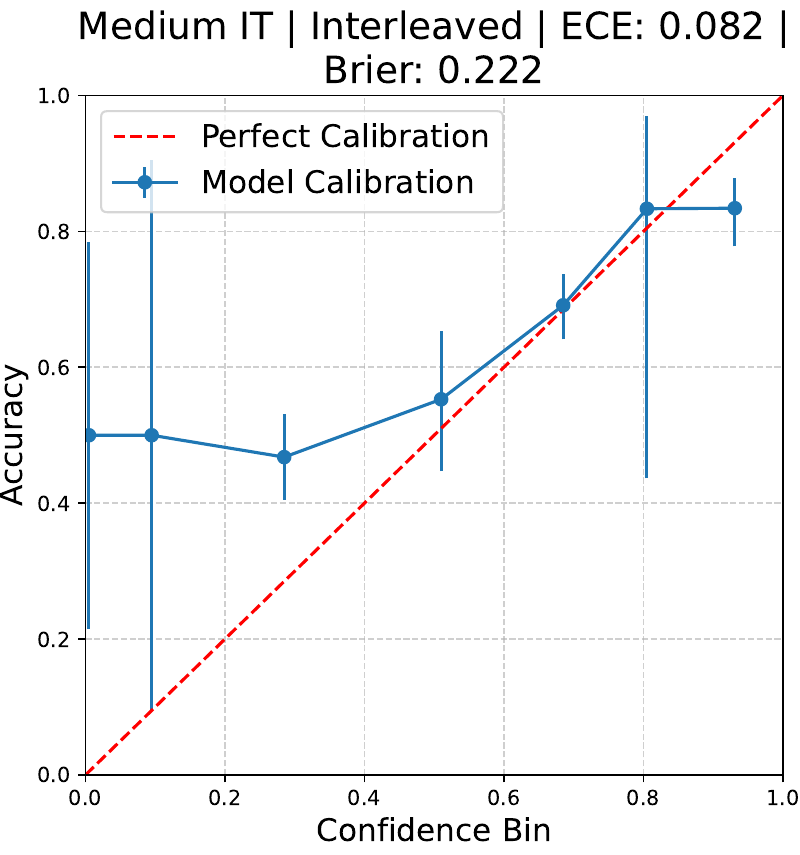}
    \caption{Prefixed Interleaved}
    \label{fig:language_pred_prob_calibration_test_set_gemini_medium_it_ambig_qa_prefixed_interleaved}
\end{subfigure}

\begin{subfigure}[b]{0.45\textwidth}
    \includegraphics[width=\textwidth]{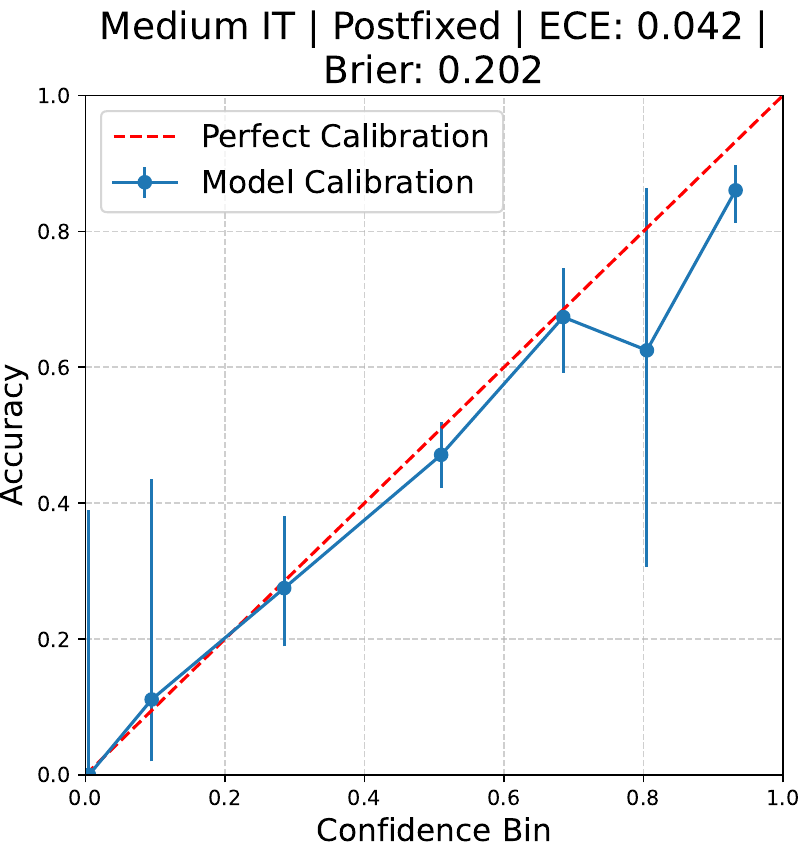}
    \caption{Postfixed}
    \label{fig:language_pred_prob_calibration_test_set_gemini_medium_it_ambig_qa_postfixed}
\end{subfigure}\hfill

\caption{\textbf{Ambig QA Test Calibration Plots of Finetuned Model (Gemini IT Medium, In Distribution)}: The x-axis is the $p_{model}(true)$ obtained by converting the linguistic uncertainty expression, in the prediction of the finetuned model, to a float (shown here as \texttt{Confidence Bin}) and y-axis is probability of that prediction being actually correct (shown here as \texttt{Accuracy}). No post-processing is done on the $p_{model}(true)$. Expected Calibration Error (ECE) and Brier Score are reported at the top of each plot.}
\label{fig:language_pred_prob_calibration_test_set_gemini_it_medium_ambig_qa}
\end{figure*}


\begin{figure*}[htp]
\captionsetup[subfigure]{justification=centering}
\centering
\begin{subfigure}[b]{0.45\textwidth}
    \includegraphics[width=\textwidth]{Figures/finetuned/language_pred_prob_calibration_test_set_gemini_small_truthful_qa_prefixed.pdf}
    \caption{Prefixed}
    \label{fig:language_pred_prob_calibration_test_set_gemini_small_truthful_qa_prefixed}
\end{subfigure}\hfill
\begin{subfigure}[b]{0.45\textwidth}
    \includegraphics[width=\textwidth]{Figures/finetuned/language_pred_prob_calibration_test_set_gemini_small_truthful_qa_prefixed_interleaved.pdf}
    \caption{Prefixed Interleaved}
    \label{fig:language_pred_prob_calibration_test_set_gemini_small_truthful_qa_prefixed_interleaved}
\end{subfigure}

\begin{subfigure}[b]{0.45\textwidth}
    \includegraphics[width=\textwidth]{Figures/finetuned/language_pred_prob_calibration_test_set_gemini_small_truthful_qa_postfixed.pdf}
    \caption{Postfixed}
    \label{fig:language_pred_prob_calibration_test_set_gemini_small_truthful_qa_postfixed}
\end{subfigure}\hfill

\caption{\textbf{Truthful QA Test Calibration Plots of Finetuned Model (Gemini Small, In Distribution)}: The x-axis is the $p_{model}(true)$ obtained by converting the linguistic uncertainty expression, in the prediction of the finetuned model, to a float (shown here as \texttt{Confidence Bin}) and y-axis is probability of that prediction being actually correct (shown here as \texttt{Accuracy}). No post-processing is done on the $p_{model}(true)$. Expected Calibration Error (ECE) and Brier Score are reported at the top of each plot.}
\label{fig:language_pred_prob_calibration_test_set_gemini_small_truthful_qa}
\end{figure*}

\begin{figure*}[htp]
\captionsetup[subfigure]{justification=centering}
\centering
\begin{subfigure}[b]{0.45\textwidth}
    \includegraphics[width=\textwidth]{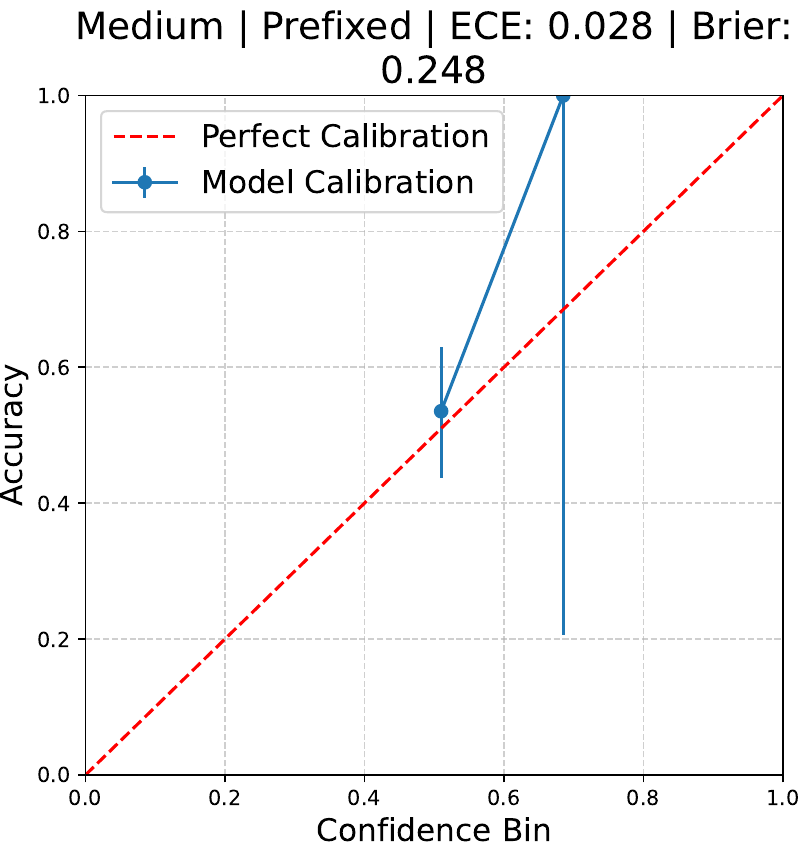}
    \caption{Prefixed}
    \label{fig:language_pred_prob_calibration_test_set_gemini_medium_truthful_qa_prefixed}
\end{subfigure}\hfill
\begin{subfigure}[b]{0.45\textwidth}
    \includegraphics[width=\textwidth]{Figures/finetuned/language_pred_prob_calibration_test_set_gemini_medium_truthful_qa_prefixed_interleaved.pdf}
    \caption{Prefixed Interleaved}
    \label{fig:language_pred_prob_calibration_test_set_gemini_medium_truthful_qa_prefixed_interleaved}
\end{subfigure}

\begin{subfigure}[b]{0.45\textwidth}
    \includegraphics[width=\textwidth]{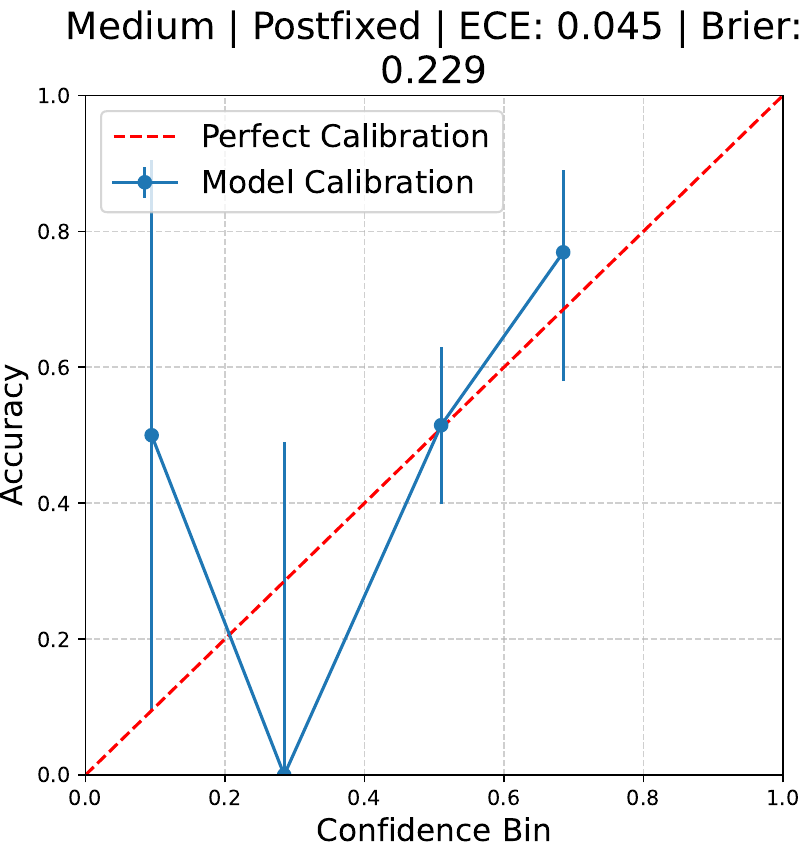}
    \caption{Postfixed}
    \label{fig:language_pred_prob_calibration_test_set_gemini_medium_truthful_qa_postfixed}
\end{subfigure}\hfill

\caption{\textbf{Truthful QA Test Calibration Plots of Finetuned Model (Gemini Medium, In Distribution)}: The x-axis is the $p_{model}(true)$ obtained by converting the linguistic uncertainty expression, in the prediction of the finetuned model, to a float (shown here as \texttt{Confidence Bin}) and y-axis is probability of that prediction being actually correct (shown here as \texttt{Accuracy}). No post-processing is done on the $p_{model}(true)$. Expected Calibration Error (ECE) and Brier Score are reported at the top of each plot.}
\label{fig:language_pred_prob_calibration_test_set_gemini_medium_truthful_qa}
\end{figure*}

\begin{figure*}[htp]
\captionsetup[subfigure]{justification=centering}
\centering
\begin{subfigure}[b]{0.45\textwidth}
    \includegraphics[width=\textwidth]{Figures/finetuned/language_pred_prob_calibration_test_set_gemini_small_it_truthful_qa_prefixed.pdf}
    \caption{Prefixed}
    \label{fig:language_pred_prob_calibration_test_set_gemini_small_it_truthful_qa_prefixed}
\end{subfigure}\hfill
\begin{subfigure}[b]{0.45\textwidth}
    \includegraphics[width=\textwidth]{Figures/finetuned/language_pred_prob_calibration_test_set_gemini_small_it_truthful_qa_prefixed_interleaved.pdf}
    \caption{Prefixed Interleaved}
    \label{fig:language_pred_prob_calibration_test_set_gemini_small_it_truthful_qa_prefixed_interleaved}
\end{subfigure}

\begin{subfigure}[b]{0.45\textwidth}
    \includegraphics[width=\textwidth]{Figures/finetuned/language_pred_prob_calibration_test_set_gemini_small_it_truthful_qa_postfixed.pdf}
    \caption{Postfixed}
    \label{fig:language_pred_prob_calibration_test_set_gemini_small_it_truthful_qa_postfixed}
\end{subfigure}\hfill

\caption{\textbf{Truthful QA Test Calibration Plots of Finetuned Model (Gemini IT Small, In Distribution)}: The x-axis is the $p_{model}(true)$ obtained by converting the linguistic uncertainty expression, in the prediction of the finetuned model, to a float (shown here as \texttt{Confidence Bin}) and y-axis is probability of that prediction being actually correct (shown here as \texttt{Accuracy}). No post-processing is done on the $p_{model}(true)$. Expected Calibration Error (ECE) and Brier Score are reported at the top of each plot.}
\label{fig:language_pred_prob_calibration_test_set_gemini_it_small_truthful_qa}
\end{figure*}

\begin{figure*}[htp]
\captionsetup[subfigure]{justification=centering}
\centering
\begin{subfigure}[b]{0.45\textwidth}
    \includegraphics[width=\textwidth]{Figures/finetuned/language_pred_prob_calibration_test_set_gemini_medium_it_truthful_qa_prefixed.pdf}
    \caption{Prefixed}
    \label{fig:language_pred_prob_calibration_test_set_gemini_medium_it_truthful_qa_prefixed}
\end{subfigure}\hfill
\begin{subfigure}[b]{0.45\textwidth}
    \includegraphics[width=\textwidth]{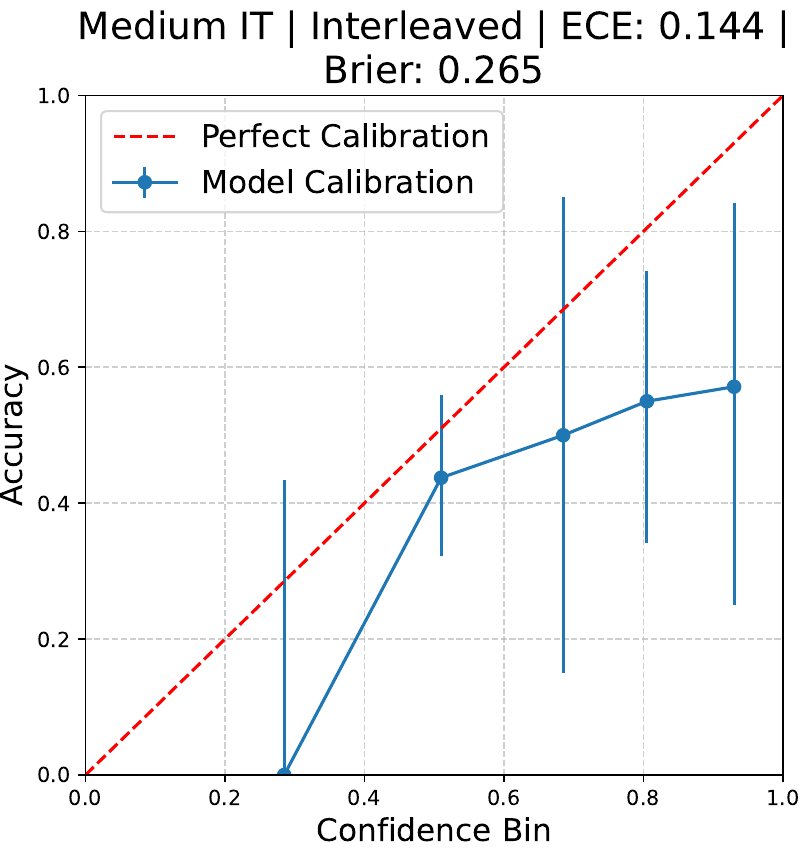}
    \caption{Prefixed Interleaved}
    \label{fig:language_pred_prob_calibration_test_set_gemini_medium_it_truthful_qa_prefixed_interleaved}
\end{subfigure}

\begin{subfigure}[b]{0.45\textwidth}
    \includegraphics[width=\textwidth]{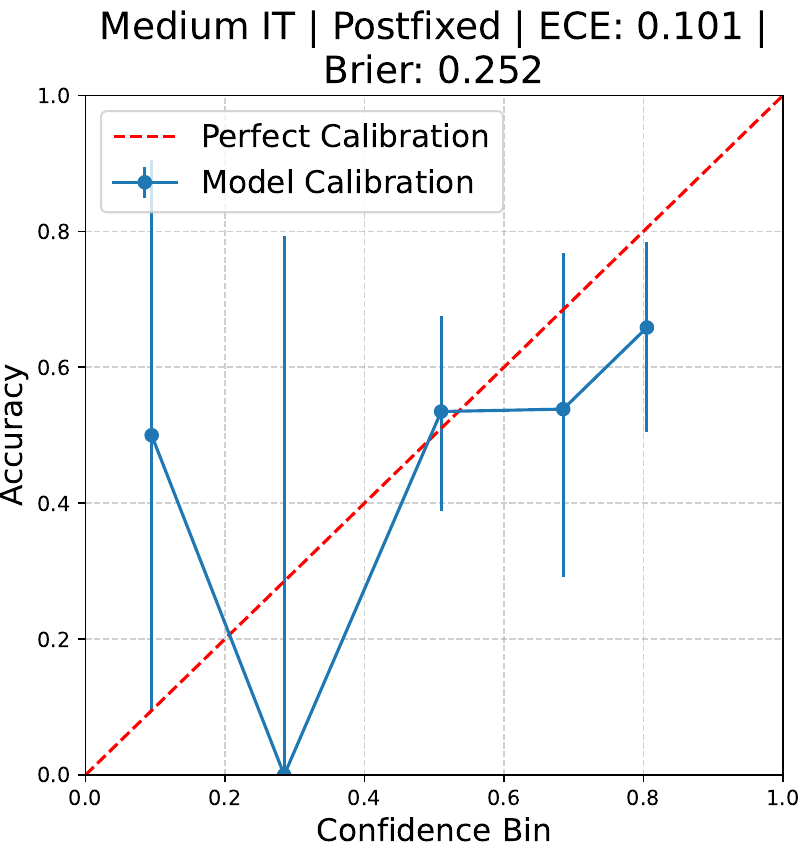}
    \caption{Postfixed}
    \label{fig:language_pred_prob_calibration_test_set_gemini_medium_it_truthful_qa_postfixed}
\end{subfigure}\hfill

\caption{\textbf{Truthful QA Test Calibration Plots of Finetuned Model (Gemini IT Medium, In Distribution)}: The x-axis is the $p_{model}(true)$ obtained by converting the linguistic uncertainty expression, in the prediction of the finetuned model, to a float (shown here as \texttt{Confidence Bin}) and y-axis is probability of that prediction being actually correct (shown here as \texttt{Accuracy}). No post-processing is done on the $p_{model}(true)$. Expected Calibration Error (ECE) and Brier Score are reported at the top of each plot.}
\label{fig:language_pred_prob_calibration_test_set_gemini_it_medium_truthful_qa}
\end{figure*}

\end{document}